\pdfoutput=1
\documentclass[journal]{IEEEtran}
\usepackage[utf8]{inputenc}
\usepackage{amsmath}
\usepackage{amssymb}
\usepackage{booktabs}
\usepackage{multirow}
\usepackage{makecell}
\usepackage{longtable}
\usepackage{graphicx}
\usepackage{tikz}
\usetikzlibrary{arrows.meta}
\usepackage{colortbl}
\definecolor{ukblue}{HTML}{002B88}
\usepackage{hyperref}

\hypersetup{
  colorlinks=true,
  linkcolor=ukblue,
  citecolor=ukblue,
  urlcolor=ukblue!85,
  pdftitle={DALE-CT: Depth-Aware 2D Slice Encoders Learn an Anatomical World Model of Chest CT},
  pdfauthor={Evan W. Damron, Mahmut S. Gokmen, Mitchell A. Klusty, Caroline N. Leach, Emily B. Collier, V. K. Cody Bumgardner},
}
\usepackage{cite}

\title{DALE-CT: Depth-Aware 2D Slice Encoders Learn an Anatomical World Model of Chest CT}
\author{Evan~W.~Damron, Mahmut~S.~Gokmen, Mitchell~A.~Klusty, Caroline~N.~Leach, Emily~B.~Collier, V.~K.~Cody~Bumgardner\\
        \textit{College of Medicine Office of Research, University of Kentucky, Lexington, KY 40506 USA} \\}

\begin{document}

\maketitle

\begin{abstract}
Chest CT is among the highest-volume imaging exams in medicine, yet the expert voxel-level annotations needed to train assistive models remain scarce and costly, motivating encoders that learn directly from unlabeled scans. We present DALE-CT, a family of 2D slice-based Vision Transformers trained entirely from scratch on chest CT with the heuristics-free LeJEPA objective. We introduce depth-aware slab sampling, which draws self-supervised views from across a physical $z$-axis slab rather than a single slice, implicitly tasking the 2D encoder with representing how anatomy changes between neighboring slices. The frozen representations trace each scan as a smooth anatomical trajectory, recover cranio-caudal slice ordering without labels, and distinguish slices by the anatomy they contain rather than by position alone. This anatomical world model emerges without any 3D or positional supervision, and an otherwise-identical encoder trained on slices in isolation never develops it. An attribution study finds the geometry costs nothing in classification accuracy, which stays robust under $8\times$ slice subsampling, and yields smoother organ coverage at some cost to boundary sharpness. Building on this backbone, we introduce dense auxiliary supervision into the pretraining objective, using anatomical and abnormality masks to supervise patch and slice tokens alongside the self-supervised loss, and we compare the resulting variants against a DINOv2 baseline continually pretrained on CT-RATE from natural-image weights. Among nine public and in-house models evaluated under the same protocol, DALE-CT-2S is the strongest 2D model in-domain, reaching 0.825 Macro AUROC on CT-RATE, within 0.024 of COLIPRI-CRM and without any text supervision. We subsequently scale the supervision-free configuration to a $\sim$287k-scan multi-source pool, to our knowledge the largest chest-CT pretraining corpus reported to date. The resulting DALE-CT-0-L achieves the best external-transfer point estimates of any 2D backbone evaluated, and we release it as our recommended backbone together with every configuration in the study, all training code, and all evaluation scripts.
\end{abstract}


\section{Introduction}

The widespread use of Computed Tomography (CT) has created an urgent need for automated diagnostic assistance systems. Adapting deep learning architectures to medical imaging is challenging because CT volumes contain severe class imbalances and complex anatomical geometries, and because expert annotations---above all voxel-level 3D annotations---remain scarce and costly to produce, motivating self-supervised learning (SSL) from unlabeled scans. Recent releases of large-scale datasets, such as CT-RATE \cite{ct-rate}, offer unprecedented scale for developing foundational models designed to address these complexities.

Currently, the standard approach relies on native 3D architectures backed by multi-objective pretraining. Models like the COLIPRI encoder family \cite{colipri} combine vision-language pretraining with image-only self-supervised learning to achieve exceptional performance. However, relying heavily on textual supervision can limit the computed visual features to the specific observations explicitly documented by radiologists \cite{rad_dino}. Because not all visual details or incidental findings are captured in standard reports, this textual bottleneck can lead to a collapse of representations at the expense of comprehensive image understanding. By contrast, isolating representation learning within a vision-only encoder allows the model to learn an exhaustive baseline of visual features first, without being constrained by report completeness. Once this robust spatial understanding is established, the unconstrained vision encoder can be grafted to a language model after the fact to narrow in on specific abnormalities—an approach that has proven highly effective, achieving state-of-the-art results in generative 3D CT tasks like Ker-VLJEPA-3B \cite{bumgardner}. Native 3D architectures also fix their input geometry, so variable-length clinical scans must first be resampled to a common grid; a slice-based encoder instead ingests every scan at its native resolution and slice count.

To address this, we systematically evaluate modern 2D self-supervised learning applied to 3D CT data. We train our foundational model family, DALE-CT (Depth-Aware Latent-Euclidean Computed Tomography), entirely from scratch with the Latent-Euclidean Joint-Embedding Predictive Architecture (LeJEPA) \cite{lejepa}---a theoretically grounded, heuristics-free objective---and benchmark it against our own domain-specific continual pretraining of DINOv2 from natural-image weights. To anatomically anchor the representations, we add an optional dense auxiliary supervision head consuming soft, fractional labels derived from TotalSegmentator \cite{totalseg} and ReXGroundingCT \cite{rex} masks. We evaluate three variants trained from scratch on CT-RATE: DALE-CT-0 (no auxiliary supervision; depth-aware LeJEPA alone), DALE-CT-1S-v2 (auto-generated TotalSegmentator anatomical labels, with ReXGroundingCT held inert), and DALE-CT-2S (both TotalSegmentator and ReXGroundingCT abnormality labels). Finally, we train DALE-CT-0-L as our strongest general-purpose release: it scales the no-auxiliary recipe to a $\sim$287k-scan multi-source chest-CT corpus---to our knowledge the largest chest-CT pretraining corpus reported to date.

Our central finding is that DALE-CT functions as an anatomical world model. The LeJEPA objective belongs to the Joint-Embedding Predictive Architecture family, proposed as a foundation for world models that predict latent states from one another \cite{lejepa,leworldmodel}. By sampling crops across a continuous physical $z$-axis slab, our depth-aware encoder is implicitly tasked with representing the anatomical transition between adjacent slices, yielding a latent space that encodes volumetric structure. We show that this structure is directly recoverable: frozen DALE-CT representations trace a smooth, continuous trajectory through each volume, recover cranio-caudal slice ordering without positional labels, and distinguish slices by the anatomy they contain rather than by where they sit---volumetric structure that encoders trained on slices in isolation demonstrably lack. Crucially, we then subject this finding to an attribution study: depth-aware training leaves classification accuracy unchanged at every slice spacing and trades boundary sharpness for smoothness in organ localization. DALE-CT is thus a structurally grounded, text-free anatomical world model for CT, and we report precisely what that structure does and does not buy.

The released models are, at the same time, highly competitive classifiers. In a benchmark where we probe every model ourselves under a single protocol on shared splits, our 2D models outperform established native 3D architectures such as Merlin \cite{merlin} and CT-FM \cite{ct-fm} on in-domain multi-abnormality classification---despite those models relying on textual supervision or larger pretraining corpora---and remain competitive with the state-of-the-art 3D model COLIPRI-CRM \cite{colipri} while bypassing the text supervision on which it relies. Externally, the scaled DALE-CT-0-L achieves the best 2D transfer point estimates, and we release it as the recommended general-purpose backbone.

\subsection{Contributions}
\begin{itemize}
    \item \textbf{DALE-CT model family.} Four 2D ViT-Large CT encoders trained from scratch with depth-aware LeJEPA: DALE-CT-0 (pure SSL), DALE-CT-1S-v2 (dense anatomical supervision), DALE-CT-2S (dense anatomical + abnormality supervision), and DALE-CT-0-L (pure SSL at $\sim$287k-scan scale).
    \item \textbf{An anatomical world model from 2D slices.} We show that depth-aware slab sampling induces frozen representations that trace smooth, continuous anatomical trajectories, recover cranio-caudal slice ordering without labels, and distinguish slices by anatomical content rather than position alone; a latent-continuity measure isolates the mechanism.
    \item \textbf{An attribution study of the geometry.} Two frozen-task probes on the matched sampling pair show that depth-aware training leaves classification accuracy unchanged, maintaining robustness under $8\times$ slice subsampling, and trades boundary sharpness for smoothness in organ-span localization.
    \item \textbf{A controlled study of dense auxiliary supervision.} A patch-size- and budget-controlled ladder (0 $\to$ 1S-v2 $\to$ 2S) isolates what anatomical and abnormality mask supervision each contribute: consistent in-domain AUROC gains, with the abnormality head's gain not carrying over to external transfer.
    \item \textbf{A head-to-head frozen-probe benchmark.} We evaluate nine public and in-house CT foundation models under one linear-probing multiple-instance-learning (MIL) protocol, on shared splits, with bootstrap confidence intervals---replacing the field's common practice of comparing self-reported numbers across incompatible protocols.
    \item \textbf{Scale versus supervision.} Scaling supervision-free pretraining to the largest reported chest-CT corpus matches anatomically supervised pretraining in-domain, achieves the best 2D external-transfer point estimates, and preserves the anatomical world model.
    \item \textbf{Open release.} All training code, evaluation scripts, and model weights are public, with every DALE-CT variant loadable directly from the Hugging Face Hub.
\end{itemize}

\section{Related Work}

\subsection{Self-Supervised Learning: DINOv2 and LeJEPA}

SSL has become a foundational paradigm for extracting robust representations from medical imaging without the bottleneck of exhaustive manual annotations. Broadly, modern vision SSL strategies diverge into two primary paradigms: Masked Image Modeling (MIM) and invariance-based learning. MIM architectures, such as Masked Autoencoders \cite{mae}, aim to reconstruct missing pixel or token information from partial observations, effectively forcing the model to understand local spatial contexts. Conversely, invariance-based methods learn by constraining the embedding space such that augmented views of the same underlying image map to similar representations.

The fundamental challenge in invariance-based learning is preventing representation collapse, where the network maps all inputs to a constant trivial vector. To counter this, early architectures relied on contrastive learning paradigms, utilizing negative sample pairs to explicitly push dissimilar representations apart within the latent space (e.g., SimCLR \cite{simclr}, MoCo \cite{moco}). While effective, contrastive methods scale poorly due to their reliance on massive batch sizes or large memory banks to maintain sufficient negative samples.

This bottleneck spurred the development of non-contrastive invariance-based methods. Instead of negative pairs, these frameworks avoid collapse through architectural heuristics—such as asymmetric teacher--student networks and exponential moving averages (e.g., BYOL \cite{byol}, DINO \cite{dino})—or through explicit statistical regularization of the latent space (e.g., VICReg \cite{vicreg}). A prominent architecture driving this non-contrastive shift is DINOv2 \cite{dinov2}, which combines the DINO image-level self-distillation objective \cite{dino} with the iBOT masked-image-modeling objective \cite{ibot} and has demonstrated exceptional capacity for learning dense, versatile features. In radiology, DINOv2-based models are increasingly utilized to capture complex anatomical geometries and abnormalities \cite{tapct} \cite{evaldinov2rad} \cite{rad_dino}.

However, traditional non-contrastive frameworks like DINO rely heavily on the aforementioned complex architectural heuristics to stabilize training, introducing significant hyperparameter and computational overhead. To address this, we utilize the Latent-Euclidean Joint-Embedding Predictive Architecture (LeJEPA) \cite{lejepa}. It entirely replaces heuristic stabilization with Sketched Isotropic Gaussian Regularization (SIGReg). By explicitly constraining the latent space to an optimal isotropic Gaussian distribution, LeJEPA ensures uniform feature coverage and achieves linear scalability, making it inherently stable and particularly well-suited for high-dimensional volumetric data.

The world-model perspective frames our analysis of these representations. A world model is a learned model of environment structure that supports predicting states the agent has not observed, originally formulated for control, where a recurrent latent dynamics model enables planning and imagination \cite{worldmodels}. LeCun's position paper places joint-embedding predictive architectures at the center of this program, arguing that world models should predict in representation space rather than pixel space \cite{lecun_path}, and subsequent work instantiates the idea without actions, as passive latent prediction over video frames \cite{vjepa} and end-to-end from pixels \cite{leworldmodel}. Recent analysis asks when LeJEPA training itself yields such a model \cite{lejepa_worldmodel_when}. We use the term in this passive sense, with the cranio-caudal axis of a CT volume playing the role the temporal axis plays in video: slices are successive states, and depth-aware training asks the encoder to relate them. To our knowledge, no prior CT foundation model frames its frozen representations as an anatomical world model.

Furthermore, recent advancements in medical vision demonstrate the critical importance of explicitly guiding self-supervised models toward clinically relevant regions. For instance, DINO-LG \cite{dinolg} is a label-guided extension of the DINO architecture that incorporates targeted augmentation on annotated calcified regions during self-supervised pretraining. By probabilistically centering crops on areas of interest, their label-guided approach improves feature discrimination for small, sparse calcifications, significantly reducing both false-negative and false-positive rates compared to standard unguided pretraining. This precedent strongly motivates our own label-guided cropping strategy, ensuring that the self-supervised representations inherently prioritize active pathology and anatomical tissue over background space.

\subsection{3D Foundation Models and Vision-Language Architectures}
The landscape of volumetric medical imaging has been heavily influenced by 3D Foundation Models and Vision-Language Models (VLMs). \textbf{CT-CLIP} \cite{ct-rate} pioneered contrastive language-image pretraining for 3D chest CTs by pairing 25,692 scans with textual reports to enable multi-abnormality detection and case retrieval without task-specific training. In contrast, \textbf{CT-FM} \cite{ct-fm} utilizes label-agnostic contrastive learning across a massive corpus of ${\sim}$148,000 scans, demonstrating strong anatomical clustering and retrieval capabilities relying solely on 3D visual data. To overcome the limitations of short text reports, \textbf{Merlin} \cite{merlin} integrates both unstructured reports and structured Electronic Health Records (EHR), yielding robust zero-shot findings and chronic disease prediction. \textbf{COLIPRI} \cite{colipri} further addresses data scarcity by combining image-only pretraining with a report generation objective, leveraging both paired and unpaired data to achieve state-of-the-art zero-shot and classification probing results. Other architectures, such as \textbf{U-VLM} \cite{uvlm}, focus on hierarchical routing, injecting multi-scale visual features from a 3D U-Net directly into corresponding language model layers after a progressive training pipeline. While highly performant, these models universally impose rigid 3D input constraints.

\subsection{2D Volumetric CT Representation and Sequence Aggregation}
While native 3D architectures often demand substantial computational overhead, 2D slice-based models provide a highly efficient foundation for volumetric representation. To process these 2D slices effectively, we employ Vision Transformers (ViTs), which divide an image into a grid of non-overlapping sub-regions called patches and "transform" them into \textbf{patch tokens}, which are high-dimensional vectors. These patch tokens preserve fine-grained, localized spatial information. A learnable global summary vector, known as the \textbf{CLS token} (classification token), is prepended to the sequence to aggregate macroscopic context for the entire slice. Throughout this paper, the term \textbf{dense} refers to operations or supervision mechanisms applied simultaneously across all localized patch tokens—rather than just the global CLS token—enabling spatially granular feature extraction.

The success of a 2D-to-3D pipeline is heavily dependent on the sophistication of the aggregation mechanism used to build a cohesive clinical context out of these slice embeddings. Recent methodologies suggest that explicitly routing 2D features through physical and anatomical constraints is essential for bridging this dimensional gap. For example, ORACLE-CT \cite{oracle-ct} demonstrated that an organ-aware pooling head injects critical spatial guidance for predicting localized diseases. Advancing this paradigm, the JANUS architecture \cite{janus} bounds these localized representations by conditioning visual embeddings on quantitative macro-radiomic priors via a multiplicative gate. 

Similarly, Ker-VLJEPA-3B \cite{bumgardner} demonstrates the scaling potential of 2D representations within complex generative pipelines. To achieve state-of-the-art results in 3D CT report generation (0.429 Macro F1 on the CT-RATE benchmark), the authors utilize our DALE-CT-1S backbone coupled with zone-constrained cross-attention. This architecture effectively bridges the modality gap by spatially compressing variable-length sequences of 2D embeddings into fixed tokens for a large language model.

Our work serves as a quantitative benchmark for the foundational representations that power these paradigms. By rigorously evaluating these backbones prior to complex downstream aggregation, we establish the baseline semantic capacity of features learned during self-supervised pretraining.

\section{Methodology}
\begin{figure*}[tp]
    \centering
    \resizebox{\textwidth}{!}{\input{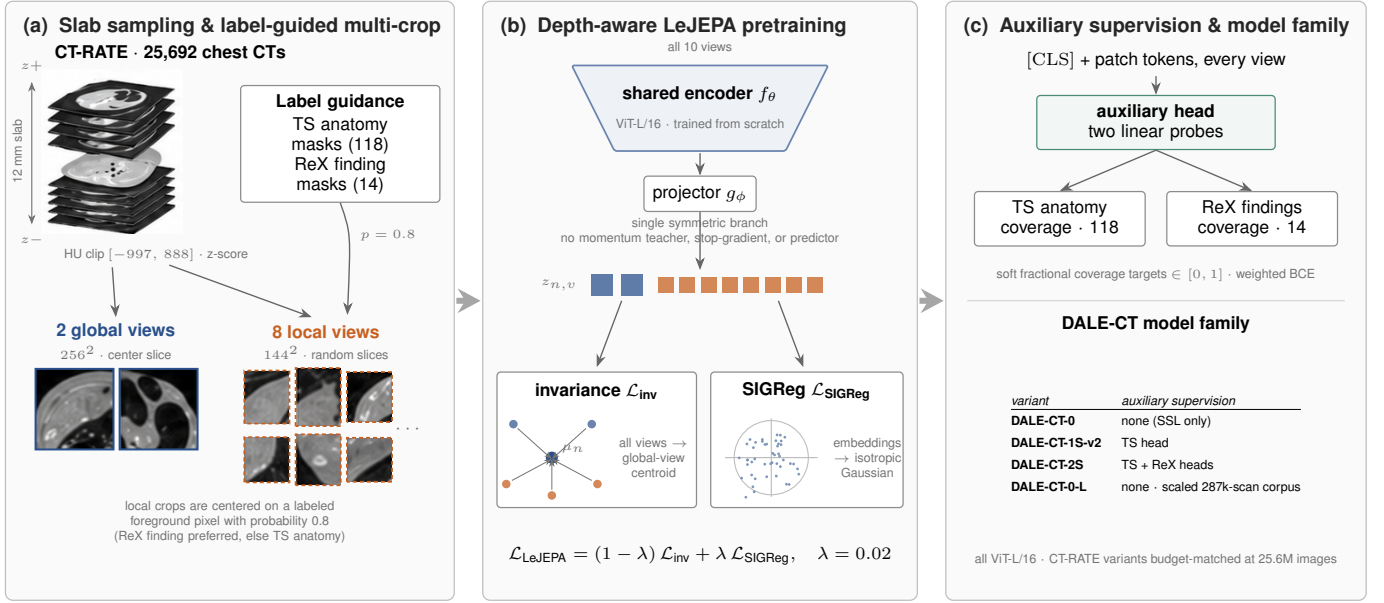}}
    \caption{Overview of the DALE-CT pretraining framework. \textbf{(a)} Each scan is grouped into continuous 12~mm slabs; every training sample draws two global views from the slab's center slice and eight local views from randomly selected slices, so the multi-crop set spans the slab's physical depth. Local crops are label-guided: with probability 0.8 they are centered on a ReXGroundingCT (ReX) finding when present, else on TotalSegmentator (TS) anatomy. \textbf{(b)} All ten views pass through a single shared ViT-L/16 encoder $f_\theta$ and projector $g_\phi$---no momentum teacher, stop-gradient, or predictor---optimized with the multi-crop invariance loss $\mathcal{L}_{\text{inv}}$, which pulls every view toward the centroid of the global views, alongside Sketched Isotropic Gaussian Regularization (SIGReg), which drives the embedding distribution toward an isotropic Gaussian and prevents collapse structurally. \textbf{(c)} An optional dense auxiliary head supervises the \texttt{[CLS]} and patch tokens of every view with soft fractional-coverage targets over the 118 TS and 14 ReX classes, yielding the DALE-CT model family.}
    \label{fig:lejepa_architecture}
\end{figure*}

\subsection{Dataset Preparation}

Our CT-RATE pretraining cohort~\cite{ct-rate} comprises 25,692 chest CTs. We convert raw NIfTI volumes to Hounsfield units (HU) and group axial slices into continuous 12~mm slabs computed from each scan's $z$-spacing. The slab thickness is chosen for robustness to the cohort's variable spacing: at the coarsest 6~mm inter-slice spacing, a 12~mm slab still spans at least three slices, so every volume contributes cross-slice context. HU values are clipped to $[-997, 888]$ and standardized ($\mu{=}{-}142$, $\sigma{=}361$); following~\cite{tapct}, the clip bounds are the 0.5/99.5 percentiles of foreground voxels and the moments their mean and standard deviation, estimated from 1,000 CT-RATE scans. (The scaled DALE-CT-0-L variant replaces this corpus with a larger multi-source pool; see Section~\ref{sec:large_corpus}.)

For dense spatial supervision we draw on two complementary annotation sources. TotalSegmentator (TS)~\cite{totalseg} is an automatic segmentation model that labels 117 anatomical structures (plus a background class); we run it on every CT-RATE scan to obtain per-structure masks. ReXGroundingCT (ReX)~\cite{rex} augments 3,142 CT-RATE scans with manually drawn, radiologist-validated 3D segmentation masks for 14 lung and pleural finding categories grounded in the free-text reports. Both mask sets are downsampled to $128 \times 128$---sufficient for the dense auxiliary supervision and label-guided cropping detailed in Sections~\ref{sec:aux_supervision} and~\ref{sec:augmentation}, at reduced dataloading cost.

\subsection{LeJEPA Formulation and Auxiliary Supervision}
\label{sec:aux_supervision}
Our foundational models are initialized from scratch and trained with LeJEPA~\cite{lejepa}, which pairs a multi-crop invariance objective with SIGReg. Unlike teacher--student frameworks, LeJEPA routes every view through a single symmetric encoder $f_\theta$ followed by a projection head $g_\phi$, with no momentum encoder, predictor, or stop-gradient. Although the models remain native 2D-slice encoders, the multi-crop views are sampled across the depth of a 3D slab composed of multiple slices (Section~\ref{sec:augmentation}); routing physically displaced, multi-scale views through the unified encoder encourages the network to learn cross-slice anatomical consistency rather than strictly intra-slice features. Fig.~\ref{fig:lejepa_architecture} overviews the full pipeline.

Let $h_{n,v} = g_\phi(f_\theta(x_{n,v}))$ denote the projected embedding of view $v$ of image $n$, with $V_g$ global and $V_l$ local views ($V = V_g + V_l$). The invariance objective pulls all views toward the centroid of the global views,
\begin{equation}
\mu_n = \frac{1}{V_g}\sum_{v=1}^{V_g} h_{n,v}, \qquad
\mathcal{L}_{\text{inv}} = \frac{1}{V}\sum_{v'=1}^{V} \|\mu_n - h_{n,v'}\|_2^2,
\end{equation}
with no stop-gradient on $\mu_n$; gradients flow through the centroid, so the objective is symmetric in all views. (LeJEPA terms this a prediction loss, though it employs no predictor network.) On its own this objective is collapse-prone, but SIGReg replaces the usual heuristics with a distributional constraint that makes collapse non-optimal: it drives the embeddings toward a non-degenerate isotropic-Gaussian distribution, whose constant representation is not a minimum of $\mathcal{L}_{\text{inv}}$.

SIGReg constrains the \emph{distribution} of the projected embeddings: it projects them along random one-dimensional directions and penalizes the distance between each projection's empirical characteristic function and that of a standard Gaussian, driving the embedding distribution toward an isotropic Gaussian (the full formulation is given in Appendix~\ref{app:sigreg}). The total self-supervised objective is
\begin{equation}
\mathcal{L}_{\text{LeJEPA}} = (1-\lambda)\,\mathcal{L}_{\text{inv}} + \lambda\,\mathcal{L}_{\text{SIGReg}},
\end{equation}
with $\lambda = 0.02$. Because the target distribution is isotropic Gaussian with unit variance, the all-constant (collapsed) embedding does not satisfy it; SIGReg thus prevents collapse structurally rather than through architectural asymmetry.

\subsubsection{Auxiliary Supervision and Model Configurations}
To ground the embeddings in anatomy, we optionally attach a dense auxiliary head to a subset of models. All variants share a ViT-Large, patch-16 backbone, initialized from scratch, and all CT-RATE variants are trained under a matched 25.6M-image budget so that the configurations differ only in their supervision (full training configurations in Appendix~\ref{app:training}). We study three configurations:
\begin{itemize}
    \item \textbf{DALE-CT-0:} the LeJEPA objective alone, no auxiliary head ($\lambda_{\text{aux}}=0$).
    \item \textbf{DALE-CT-1S-v2:} a single-source (1S) auxiliary head over the 118 TotalSegmentator classes, applied densely to both the \texttt{[CLS]} and patch tokens; ReXGroundingCT is held inert, isolating the TS objective against the 2S model.
    \item \textbf{DALE-CT-2S:} a dual-source (2S) head over the 118 TS and 14 ReX classes jointly.
\end{itemize}
The auxiliary head is a pair of linear layers mapping the backbone feature dimension to the TS and ReX class counts. For every crop, both the global \texttt{[CLS]} token and the patch tokens are supervised, so each crop contributes a CLS-level and a patch-level prediction. Targets are dense fractional coverage: for TS, masks are one-hot over 118 classes; for ReX, binary abnormality channels are binarized per pixel and restricted to scans carrying ReX annotations. Coverage is the mean over the crop and the average-pool over the patch grid, yielding continuous targets in $[0,1]$ for both sources. The loss is BCE-with-logits with positive class weights to counter imbalance: TS weights are per-class (background and out-of-distribution anatomies at 1.0, foreground up to 1000), ReX weights are uniformly 800.

Combining CLS and patch, and TS and ReX, the 2S auxiliary loss is
\begin{equation}
\begin{aligned}
\mathcal{L}_{\text{Aux\_2S}} &= \tfrac{1}{2}\bigl(\mathcal{L}_{\text{cls}} + \mathcal{L}_{\text{patch}}\bigr), \\
\mathcal{L}_{\text{cls}} &= \tfrac{1}{2}(\mathcal{L}_{\text{ts\_cls}} + \mathcal{L}_{\text{rex\_cls}}), \\
\mathcal{L}_{\text{patch}} &= \tfrac{1}{2}(\mathcal{L}_{\text{ts\_patch}} + \mathcal{L}_{\text{rex\_patch}}).
\end{aligned}
\end{equation}
averaged equally over the global and local crops: $\mathcal{L}_{\text{Aux\_2S}} = \tfrac{1}{2}(\mathcal{L}_{\text{Aux\_2S}}^{\text{global}} + \mathcal{L}_{\text{Aux\_2S}}^{\text{local}})$. The 1S-v2 loss omits the ReX terms, retaining the same CLS/patch and global/local averaging. The total objective is
\begin{equation}
\mathcal{L}_{\text{Total}} = \mathcal{L}_{\text{LeJEPA}} + \lambda_{\text{aux}}\,\mathcal{L}_{\text{Aux}}, \qquad \lambda_{\text{aux}} = 0.1.
\end{equation}

\subsection{Data Augmentation and Depth-Aware Sampling}
\label{sec:augmentation}

All DALE-CT variants share a common GPU-accelerated augmentation pipeline: each view is a scaled random crop of a source slice, resized to its target resolution and passed through standard photometric transforms and z-score normalization (the full transform chain is specified in Appendix~\ref{app:training}).

Each training sample is a multi-crop set of two global crops ($256{\times}256$, scale $[0.6,1.0]$) drawn from the center slice of the slab and eight local crops ($144{\times}144$, scale $[0.3,0.6]$) drawn from a randomly selected slice of the slab. Local crops are label-guided: with probability $0.80$ their center is placed on a foreground pixel of an annotation mask, falling back to TS foreground when the preferred label is absent. For the CT-RATE models the preferred guidance label is a ReX abnormality mask when present; DALE-CT-0-L disables ReX guidance entirely, as the multi-source pool lacks ReX labels outside CT-RATE, and guides local crops on TS foreground alone.

The defining feature of the pipeline is depth-aware slab sampling. Crops are drawn from a multi-slice slab rather than a single slice, so the invariance loss is computed over physically displaced, multi-scale views spanning neighboring $z$-positions. This implicitly tasks the 2D encoder with representing the anatomical transition between adjacent slices: to be invariant across the slab, the encoder must encode where in the volume each slice lies---an inductive bias for volumetric structure rather than mere data variation, which Section~\ref{sec:world_model} quantifies. The two regimes instantiate the slab differently (Table~\ref{tab:augmentation_configs} in Appendix~\ref{app:training}): the CT-RATE models sample a continuous 12~mm physical slab, while DALE-CT-0-L draws three slices from a native-isotropic zarr store (chunked per slice, $z_1$) as a non-overlapping 3-slice slab.

\subsection{Continual Pretraining of the DINOv2 Baseline}
We include DINOv2~\cite{dinov2-registers,dinov2} (the dominant paradigm in 2D self-supervised learning) as a reference arm. A ViT-Large with Registers, initialized from natural-image weights, is continually pretrained on CT-RATE via DINO-MX~\cite{dinomx}, combining DINO distillation from a momentum teacher with iBOT masked image modeling; we benchmark the result as \textit{Finetuned DINOv2}. The two paradigms differ in how they stabilize learning: DINOv2 relies on a teacher--student asymmetry with stop-gradients, whereas LeJEPA uses a symmetric joint-embedding objective with an isotropic-Gaussian regularizer. Following~\cite{cardvit}, the trained RGB patch-embedding weights are averaged to adapt the pretrained network to single-channel CT.

Crops are drawn from a 12~mm physical slab for comparability with our other arms. Each volume yields two global 224$\times$224 crops (the input resolution of the pre-initialized weights) and eight local 144$\times$144 crops, six label-guided and two random; guided crops target a ReX finding with probability 0.8 when present, otherwise a TotalSegmentator structure, as in the LeJEPA arms. The full training configuration is provided in the model card.\footnote{\url{https://huggingface.co/Kentucky-Open-Science/Finetuned-DINOv2-Chest-CT}}

\subsection{Large-Scale Chest-CT Pretraining Corpus}
\label{sec:large_corpus}

DALE-CT-0-L is built to be our strongest general-purpose release: it shares the ViT-Large backbone and depth-aware LeJEPA recipe of the CT-RATE DALE-CT-0 arm but is pretrained on a far larger corpus, scaling data rather than model capacity. The training corpus is a chest-CT-only, multi-source pool spanning 32 collections, with 287,302 scans used for training. Two cohorts dominate: the National Lung Screening Trial (NLST; ${\sim}$130k scans) and an institutional chest-CT archive, together roughly three-quarters of the training pool, with the remainder drawn from ${\sim}$30 public collections (RSNA pulmonary embolism, STOIC, 4D-Lung, and COVID cohorts among them). To our knowledge this is the largest chest-specific CT pretraining corpus reported, and comparable in scale to the largest universal CT corpora (FlexiCT, 266,227 volumes from 56 datasets~\cite{flexict}; CT-FM, ${\sim}$148,000 Imaging Data Commons scans~\cite{ct-fm}).

DALE-CT-0-L disables the auxiliary head: ReX pathology labels exist only on the CT-RATE subset of the pool, so the supervision-free configuration is the one that scales to the full corpus. Otherwise it inherits the DALE-CT-0 recipe, differing only in the slab and normalization statistics already contrasted in Table~\ref{tab:augmentation_configs} (a 3-slice native-$z_1$ slab and full-pool z-score statistics). Training runs for 3 epochs (191,357 iterations, global batch 384 on 16 H100 GPUs). The outcomes of this scaling are reported in Section~\ref{sec:scale}.

\section{An Anatomical World Model: Emergence and Attribution}
\label{sec:worldmodel_section}

We first quantify the geometric effect of depth-aware slab sampling and the anatomical world model it induces in frozen representations, then test whether this geometry translates into task performance, finding that classification is unaffected and that organ-span localization trades boundary sharpness for smoothness.

\subsection{The Geometric Impact of Depth-Aware Slab Sampling}

\begin{figure}[htbp]
    \centering
    \includegraphics[width=\columnwidth]{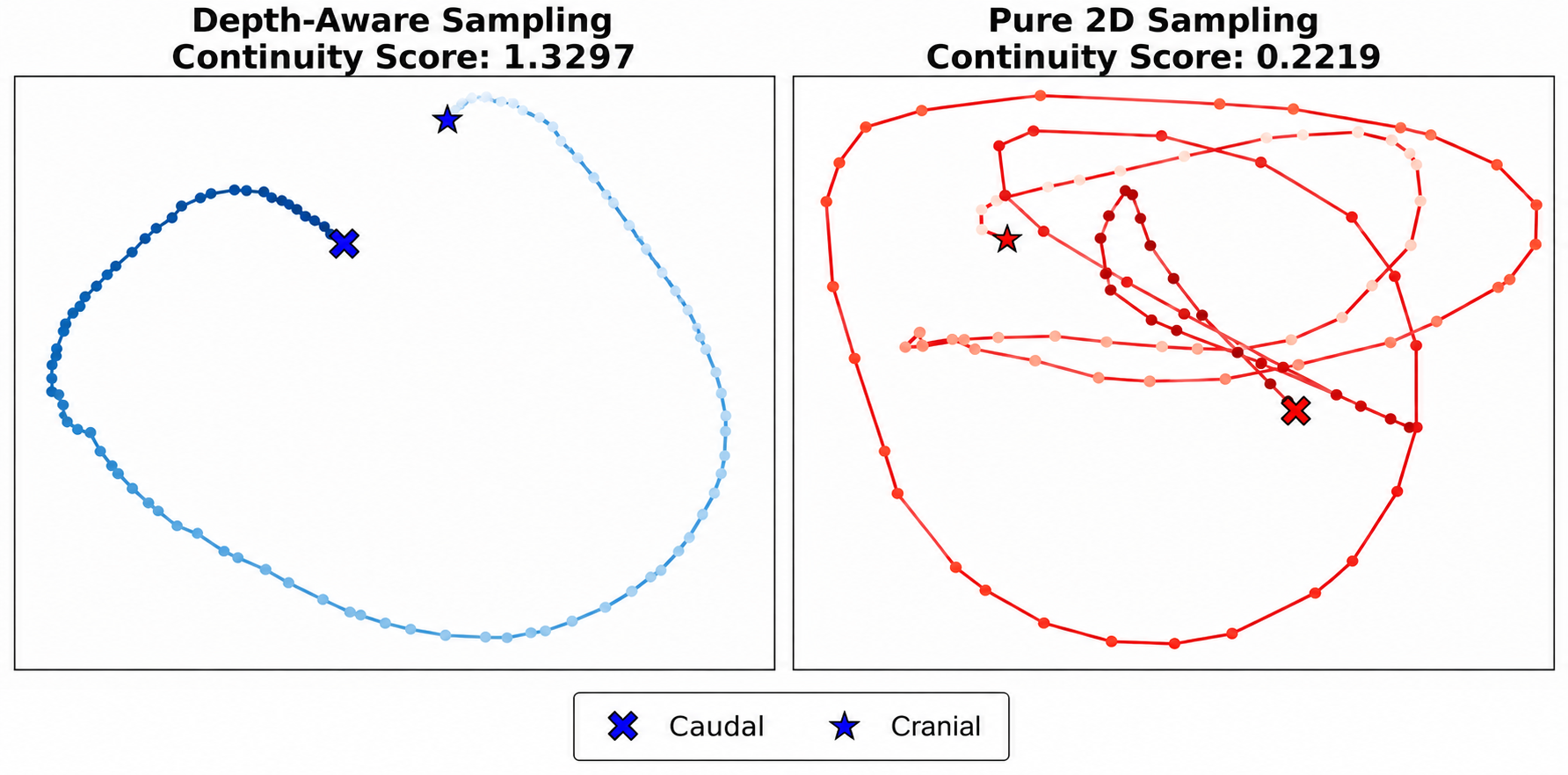}
    \caption{Comparison of Latent Continuity between depth-aware slab sampling (left) and pure-2D intra-slice sampling (right) for all slices in a representative volume (the scores shown are for the depicted volume; Table~\ref{tab:ablation_results} reports cohort means). The depth-aware model learns a smooth, continuous anatomical trajectory, while the pure-2D model exhibits tangled and disconnected representations.}
    \label{fig:slice_continuity}
\end{figure}

While our primary foundational models use a depth-aware sampling strategy---dynamically generating multi-crop views from across a continuous 12~mm physical slab---we conduct an ablation study to quantify the impact of this spatial context against a strictly 2D, intra-slice sampling approach. Using a ViT-Base architecture trained under the LeJEPA objective, we compare downstream classification efficacy (linear-probing MIL evaluation; probe details in Appendix~\ref{app:attribution}) alongside empirical measures of spatial continuity across the CT-RATE validation split. The results are summarized in Table~\ref{tab:ablation_results}.

\begin{table}[htbp]
\centering
\caption{Ablation of 2D Intra-Slice vs. Depth-Aware Slab Sampling (ViT-Base)}
\label{tab:ablation_results}
\setlength{\tabcolsep}{3pt}
\begin{tabular}{@{} l c c @{}}
\toprule
\rowcolor{ukblue!10}\textbf{Metric} & \textbf{Pure-2D} & \textbf{Depth-Aware Slab} \\
\rowcolor{ukblue!10}& \textbf{(Intra-Slice)} & \textbf{(12~mm)} \\
\midrule
\multicolumn{3}{@{}c@{}}{\textbf{Downstream Classification (Linear Probe)}} \\
\midrule
Macro AUROC & 0.7826 & \textbf{0.7838} \\
Macro AUPRC & \textbf{0.4647} & 0.4631 \\
Macro F1 & \textbf{0.4920} & 0.4872 \\
Balanced Acc. & 0.6930 & \textbf{0.6932} \\
\midrule
\multicolumn{3}{@{}c@{}}{\textbf{Latent Geometric Structure ($N=3002$)}} \\
\midrule
Latent Continuity Score $\uparrow$ & 0.4263$\pm$0.0049 & \textbf{1.6844$\pm$0.0472} \\
\bottomrule
\end{tabular}
\end{table}

As detailed in Table~\ref{tab:ablation_results}, downstream linear probing reveals that depth-aware slab sampling produces negligible differences in raw discriminative capacity compared to pure-2D training. However, evaluating the raw geometric structure of the resulting latent spaces reveals a clear representational shift.

To quantify this, we compute a Latent Continuity Score on the raw, un-pooled slice embeddings. Latent Continuity measures the efficiency and smoothness of the model's trajectory from the cranial to the caudal end of the CT volume in the high-dimensional feature space. It is calculated as the macroscopic distance (the direct cosine distance between the first and last slice) divided by the total path length (the sum of cosine distances between all consecutive slices). Intuitively, a model that perfectly understands continuous anatomical progression will traverse the latent space directly and smoothly, yielding a high continuity score. Conversely, a model lacking spatial awareness will produce an erratic, highly tangled trajectory, inflating the denominator and lowering the continuity score.

\begin{figure*}[tp]
    \centering
    \includegraphics[width=\textwidth]{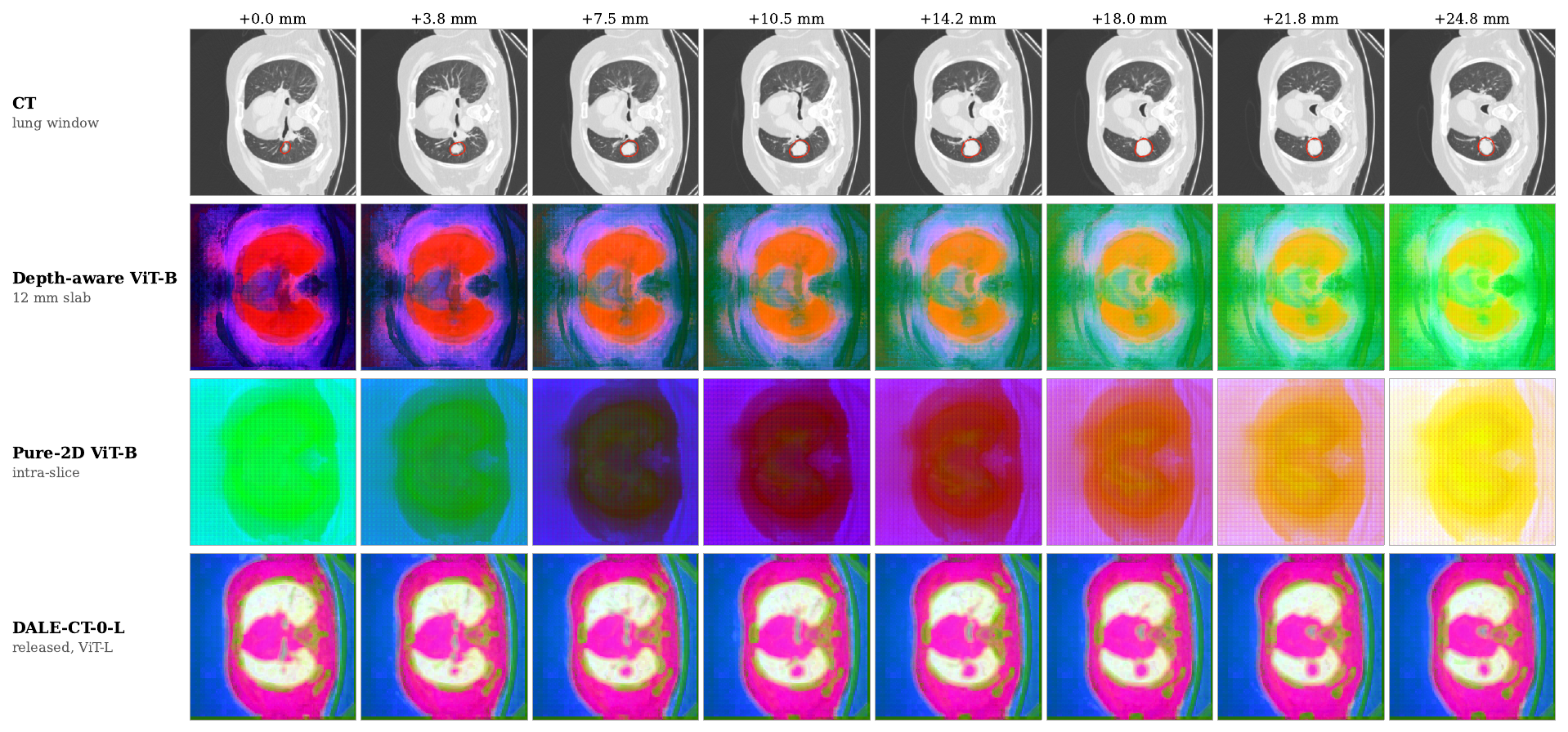}
    \caption{\textbf{Patch-level feature continuity along $z$.} Eight axial slices spanning 24.8~mm through a solitary lung mass in CT-RATE validation volume \texttt{valid\_660\_a\_2} (0.75~mm slice spacing). The ReXGroundingCT annotation for this finding reads \emph{``37~mm mass adjacent to the hilar region in the upper lobe of the left lung''}; its mask is outlined in red on the CT row. The remaining rows render the first three principal components of each frozen backbone's final-layer patch tokens as RGB. \emph{The PCA basis is fit once per backbone over all eight slices jointly and then applied to each}, so color that persists across a row denotes a feature that persists; a per-slice basis would give every panel an arbitrary basis and sign, and is shown as a control in Fig.~\ref{fig:z_continuity_perslice}. To resolve structure below the 14~px patch stride, each map is an offset-ensembled lattice: the frozen model is re-run $4\times4$ times per slice with the crop window displaced by fractions of a patch and the token grids interleaved, yielding a $148\times148$ map in which every token is a genuine model output and no learned upsampler is involved. The pure-2D map is dominated by a global component that shifts with $z$, leaving little persistent anatomical structure, while the depth-aware map retains organ structure across the same span and the released DALE-CT-0-L is stable throughout.}
    \label{fig:z_continuity}
\end{figure*}

The depth-aware model demonstrates vastly superior Latent Continuity compared to the pure-2D model (1.6844 vs.\ 0.4263).\footnote{Cosine distance ($1-\cos$) violates the triangle inequality, so this ratio can exceed $1$; an equivalent score using L2 distance on L2-normalized embeddings (a proper metric, bounded in $[0,1]$) yields $0.078$ vs.\ $0.041$, preserving the ranking.} As illustrated visually by the PCA projections in Fig.~\ref{fig:slice_continuity}, the depth-aware embeddings map the volume as a smooth, continuous trajectory. In stark contrast, the pure-2D embeddings fold over themselves into tangled, disconnected loops. Fig.~\ref{fig:z_continuity} shows the same contrast at the level of individual patch tokens rather than whole-slice embeddings: across 25~mm of a volume containing a solitary lung mass, the depth-aware backbone retains recognizable organ structure under a basis shared across slices, while the pure-2D backbone's map is dominated by a single global component that shifts with $z$. The Latent Continuity Score summarizes across 3{,}002 scans what that figure shows for one. By sampling crops across a physical $z$-axis during the self-supervised phase, the depth-aware model learns a globally coherent anatomical progression through the volume. Cross-slice sampling is thus the mechanism that injects native volumetric continuity into 2D Vision Transformers; whether that continuity translates into task performance is a separate question, which Section~\ref{sec:attribution} explores.

\subsection{Volumetric Localization from Frozen Embeddings}
\label{sec:world_model}

The smooth latent trajectory induced by depth-aware sampling (Table~\ref{tab:ablation_results}; Fig.~\ref{fig:slice_continuity}) suggests the embedding space does not merely classify slices but encodes their volumetric position. We test this directly on the same ViT-Base depth-aware and pure-2D ablation backbones, asking whether anatomical $z$-position is recoverable from frozen per-slice \texttt{[CLS]} embeddings---\emph{without any positional supervision} during pretraining.

\textbf{Slice-position regression.} We fit a ridge regression from each slice's \texttt{[CLS]} embedding to its normalized anatomical $z\in[0,1]$ (cranial~$\to$~caudal) on an 80/20 patient-level split of the 3{,}002-scan CT-RATE validation set (593 test scans). Both backbones linearly decode slice position with high fidelity (Table~\ref{tab:world_model}): depth-aware $R^2{=}0.971$ and pure-2D $R^2{=}0.974$ (rank correlation $\rho{=}0.998$ for both, mean error ${\approx}11$ slices out of several hundred). Thus, 2D encoders trained under LeJEPA spontaneously learn a cranio-caudal coordinate, whether the invariance is imposed within a single slice or across adjacent slab slices. The properties that separate the arms are the ordering and content structure that follow. The released DALE-CT-0-L extends this to production scale: trained on the $\sim$287k-scan multi-source corpus, it decodes $z$-position with comparable fidelity ($R^2{=}0.973$; Table~\ref{tab:world_model}), with bootstrap CIs that overlap the ViT-Base ablation backbones---indicating that an $11{\times}$ increase in pretraining corpus preserves the anatomical coordinate (Section~\ref{sec:scale}); its patch-level feature map is correspondingly stable across the same slices (Fig.~\ref{fig:z_continuity}, bottom row).

\textbf{Unsupervised slice-ordering recovery.} The volumetric structure is further recoverable \emph{without any labels}: sorting each scan's slices by the first principal component of their embeddings recovers the true cranio-caudal order far more faithfully for the depth-aware backbone ($|\tau|{=}0.423$ vs.\ $0.206$ for pure-2D, $|$Kendall $\tau|$, bootstrap 95\% CIs disjoint), consistent with its higher Latent Continuity (1.68 vs.\ 0.43). A local nearest-neighbor chaining variant instead favors the sharper per-slice distinctness of pure-2D features ($|\tau|{=}0.555$ vs.\ $0.487$), indicating the slab trades local $z$-sharpness for global manifold continuity.

\textbf{Anatomy beyond $z$-position.} A world model should know \emph{what} a slice shows, not merely \emph{where} it sits---and the depth-aware advantage is not merely a smoother $z$-axis: after projecting the dominant linear $z$-direction out of the frozen \texttt{[CLS]} embeddings and clustering the residuals within each scan, depth-aware features align with the TotalSegmentator organ-presence pattern far more faithfully than pure-2D ($k$-means; normalized mutual information, NMI, $0.755$ vs.\ $0.458$; adjusted Rand index, ARI, $0.662$ vs.\ $0.258$; Table~\ref{tab:world_model}; 2,000-resample bootstrap 95\% CIs disjoint; full protocol in Appendix~\ref{app:probes}). Because this is measured in $z$-residualized space, the gain cannot be attributed to $z$-position alone, indicating depth-aware slab sampling encodes organ identity that pure intra-slice training does not. Fig.~\ref{fig:z_continuity} makes the same distinction directly visible: what varies across slices in the pure-2D map is largely one global offset, leaving little residual structure once that direction is removed---which is what the $z$-residualized NMI and ARI quantify.

Together, these results show that DALE-CT's frozen representations encode an anatomical world model: a smooth volumetric coordinate, label-free global slice ordering, and anatomical content beyond position---learned from 2D slices without explicit 3D or positional supervision, and, except for position decodability itself, absent from the slice-independent control.

\begin{table}[htbp]
\centering
\caption{World-model probes on frozen \texttt{[CLS]} embeddings.}
\label{tab:world_model}
\small
\setlength{\tabcolsep}{4pt}
\resizebox{\columnwidth}{!}{%
\begin{tabular}{lccc}
\toprule
\rowcolor{ukblue!10} & \textbf{Depth-Aware} & \textbf{Pure-2D} & \textbf{DALE-CT-0-L} \\
\rowcolor{ukblue!10} & \multicolumn{2}{c}{\textit{ViT-Base ablation}} & \textit{ViT-Large, released} \\
\cmidrule(lr){2-3}\cmidrule(lr){4-4}
\midrule
$z$-regression $R^2$ & \makecell{0.971 \\ {\color{gray}\scriptsize $[0.968, 0.973]$}} & \makecell{\textbf{0.974} \\ {\color{gray}\scriptsize $[0.972, 0.977]$}} & \makecell{0.973 \\ {\color{gray}\scriptsize $[0.971, 0.976]$}} \\
\addlinespace
Order $\tau$ (PCA-1D) & \makecell{0.423 \\ {\color{gray}\scriptsize $[0.410, 0.437]$}} & \makecell{0.206 \\ {\color{gray}\scriptsize $[0.191, 0.220]$}} & \makecell{\textbf{0.559} \\ {\color{gray}\scriptsize $[0.543, 0.576]$}} \\
\addlinespace
Order $\tau$ (greedy-NN) & \makecell{0.487 \\ {\color{gray}\scriptsize $[0.461, 0.512]$}} & \makecell{\textbf{0.555} \\ {\color{gray}\scriptsize $[0.532, 0.580]$}} & \makecell{0.428 \\ {\color{gray}\scriptsize $[0.407, 0.451]$}} \\
\addlinespace
Organ NMI ($z$-resid.) & \makecell{\textbf{0.755} \\ {\color{gray}\scriptsize $[0.742, 0.769]$}} & \makecell{0.458 \\ {\color{gray}\scriptsize $[0.439, 0.479]$}} & \makecell{0.725 \\ {\color{gray}\scriptsize $[0.711, 0.738]$}} \\
\addlinespace
Organ ARI ($z$-resid.) & \makecell{\textbf{0.662} \\ {\color{gray}\scriptsize $[0.639, 0.686]$}} & \makecell{0.258 \\ {\color{gray}\scriptsize $[0.237, 0.281]$}} & \makecell{0.606 \\ {\color{gray}\scriptsize $[0.583, 0.629]$}} \\
\bottomrule
\end{tabular}}
\vspace{1ex}
\raggedright
\footnotesize \textit{Note: Columns 1--2 are the ViT-Base depth-aware vs.\ pure-2D ablation; column 3 is the released DALE-CT-0-L (ViT-Large, $\sim$287k-scan multi-source corpus). $z$-regression is supervised (ridge; 593 test scans); ordering is unsupervised ($|$Kendall $\tau|$, 2,000-resample bootstrap; 593 test scans); the Organ rows are the $z$-residualized clustering probe of Appendix~\ref{app:probes} (86 test scans). Best value per row is bolded (higher is better); gray brackets are 2,000-resample bootstrap 95\% CIs. The DALE-CT-0-L column shows the structure is preserved at pretraining scale (Section~\ref{sec:scale}).}
\end{table}

\subsection{Does the Geometry Buy Task Performance?}
\label{sec:attribution}

The preceding sections establish that depth-aware sampling produces a distinctive latent geometry. A natural inference is that this geometry is what makes the models more useful. We test the inference directly with two frozen-feature task probes on the matched ViT-Base pair, which isolates the sampling strategy at fixed capacity and corpus (full protocols in Appendix~\ref{app:attribution}). The answer is consistent: the geometry is real, but it does not translate into task performance. Classification is unchanged at every slice spacing, and organ-span localization trades boundary sharpness for smoothness.

\textbf{Organ-span localization.} We predict the contiguous slice range containing each of 12 organs from a linear coverage probe on the frozen \texttt{[CLS]} embeddings, extracting the largest connected run above a validation-tuned relative threshold (Table~\ref{tab:attribution_span}; protocol, largest-run extraction, and per-organ results in Appendix~\ref{app:attribution}, Table~\ref{tab:expf_per_organ}). Pure-2D wins: the paired span-IoU delta (depth-aware $-$ pure-2D) is negative for all 12 organs and excludes zero for 9 of 12. The smoothness prediction is confirmed but does not improve localization. The depth-aware arm produces the \emph{fewest} disconnected runs, the smoothest coverage curves exactly as the latent geometry predicts, yet the \emph{blurriest} boundaries (mean absolute error, MAE, of 9.9 vs.\ 6.5 slices). Pure-2D flickers more, but its sharper per-slice features localize boundaries better, and largest-run extraction absorbs the flicker. Regardless of attribution, both model types support frozen-feature slice-range selection for many organs, which reduces the reliance on segmentation models at inference; the released DALE-CT-0-L performs equivalently on this task.

\begin{table}[htbp]
\centering
\caption{Organ-span localization from frozen \texttt{[CLS]} probes (macro over 12 organs, 86 test scans; 95\% scan-bootstrap CIs). The released DALE-CT-0-L is shown for reference; the attribution contrast is the matched ViT-B pair, over which the best value per column is bolded.}
\label{tab:attribution_span}
\setlength{\tabcolsep}{4pt}
\begin{tabular}{lccc}
\toprule
\rowcolor{ukblue!10}\textbf{Arm} & \textbf{Span IoU} & \textbf{Boundary MAE} & \textbf{Disconnected} \\
\rowcolor{ukblue!10} & & \textbf{(slices)} & \textbf{runs} \\
\midrule
Pure-2D ViT-B & \makecell{\textbf{0.798} \\ {\color{gray}\scriptsize $[0.785, 0.811]$}} & \makecell{\textbf{6.48} \\ {\color{gray}\scriptsize $[5.74, 7.31]$}} & \makecell{2.21 \\ {\color{gray}\scriptsize $[2.06, 2.36]$}} \\
\addlinespace
Depth-aware ViT-B & \makecell{0.750 \\ {\color{gray}\scriptsize $[0.733, 0.766]$}} & \makecell{9.91 \\ {\color{gray}\scriptsize $[8.41, 11.63]$}} & \makecell{\textbf{1.78} \\ {\color{gray}\scriptsize $[1.65, 1.92]$}} \\
\addlinespace
DALE-CT-0-L & \makecell{0.786 \\ {\color{gray}\scriptsize $[0.771, 0.800]$}} & \makecell{8.33 \\ {\color{gray}\scriptsize $[7.07, 9.80]$}} & \makecell{4.29 \\ {\color{gray}\scriptsize $[3.93, 4.66]$}} \\
\bottomrule
\end{tabular}
\end{table}

\textbf{Slice-spacing robustness of classification.} Finally, we re-evaluate the exact probes behind Table~\ref{tab:ablation_results}---each arm's winning learned-attention probe and its validation-fit thresholds, unmodified---when only every $k$-th slice of each test bag is kept (Table~\ref{tab:attribution_stride}; probes trained at native spacing, same full 3{,}002-scan validation split, so the $k{=}1$ column reproduces Table~\ref{tab:ablation_results} exactly). Robustness is universal. Both arms lose at most $0.002$ macro area under the precision--recall curve (AUPRC) keeping one slice in eight, so 2D slice-MIL classification is robust to slice spacing regardless of the pretraining sampling strategy. This matters for deployment because clinical scans arrive with widely varying $z$-spacing.

\begin{table}[htbp]
\centering
\caption{Macro AUPRC under test-time slice subsampling (stride $k$), evaluated with the exact probes behind Table~\ref{tab:ablation_results} on the full validation split; the $k{=}1$ column reproduces Table~\ref{tab:ablation_results}.}
\label{tab:attribution_stride}
\setlength{\tabcolsep}{4pt}
\begin{tabular}{lccccc}
\toprule
\rowcolor{ukblue!10}\textbf{Arm} & $k{=}1$ & $k{=}2$ & $k{=}4$ & $k{=}8$ & $\Delta(1{\to}8)$ \\
\midrule
Pure-2D ViT-B & 0.4647 & 0.4646 & 0.4645 & 0.4630 & $-$0.0017 \\
Depth-aware ViT-B & 0.4631 & 0.4631 & 0.4628 & 0.4616 & $-$0.0015 \\
\bottomrule
\end{tabular}
\end{table}

\textbf{What the attribution changes.} Depth-aware sampling is the mechanism behind the world model of Section~\ref{sec:world_model}, and it again yields the smoothest curves here. But at matched capacity, our probes could not convert the geometry into task performance. Classification is unchanged at every slice spacing, and the smoothness itself costs boundary sharpness in organ-span localization. We therefore state the world-model result as what it is, a property of the representation geometry. These experiments are not proof that the geometry does nothing downstream; they show only that two simple frozen probes could not detect a benefit. State-of-the-art pipelines built on these embeddings, such as advanced MIL aggregation and report generation with Ker-VLJEPA-3B~\cite{bumgardner}, apply learned aggregation over the per-slice embedding sequence, where a smoother, better-ordered manifold could plausibly help. We leave that question to future work.

\section{Benchmark Results: Supervision, Transfer, and Scale}
\label{sec:benchmark_section}

Having established what depth-aware training changes about the representation and what it does not change about task performance, we now turn to dense auxiliary supervision. We first verify that it learns its targets, then benchmark in-domain classification, where it helps, and external transfer, where its gains do not carry over. That trade-off motivated scaling the supervision-free configuration, which closes the section.

\subsection{Auxiliary Supervision Learns Its Targets}
\label{sec:aux_verify}

\begin{table*}[t]
\centering
\caption{Auxiliary Task Linear Probing Results (414 ReX-Annotated Validation Scans, 70/10/20 Split)}
\label{tab:aux_probe_results_wide}
\begin{tabular}{lcccccccc}
\toprule
& \multicolumn{4}{c}{\textbf{ReX (14-class ML)}} & \multicolumn{4}{c}{\textbf{TS (118-class ML)}} \\
\cmidrule(lr){2-5} \cmidrule(lr){6-9}
& \multicolumn{2}{c}{\textbf{CLS Probe}} & \multicolumn{2}{c}{\textbf{Patch Probe}} & \multicolumn{2}{c}{\textbf{CLS Probe}} & \multicolumn{2}{c}{\textbf{Patch Probe}} \\
\cmidrule(lr){2-3} \cmidrule(lr){4-5} \cmidrule(lr){6-7} \cmidrule(lr){8-9}
\rowcolor{ukblue!10}\textbf{Model} & \textbf{AUROC} & \textbf{AUPRC} & \textbf{AUROC} & \textbf{AUPRC} & \textbf{AUROC} & \textbf{AUPRC} & \textbf{AUROC} & \textbf{AUPRC} \\
\midrule
Finetuned DINOv2 & 0.6636 & 0.0590 & \textbf{0.9473} & 0.0435 & 0.7683 & 0.5084 & \textbf{0.9644} & 0.4337 \\
DALE-CT-0        & 0.7457 & 0.0839 & 0.8926 & 0.0177 & 0.9151 & 0.7740 & 0.9546 & 0.2938 \\
DALE-CT-0-L      & 0.7241 & 0.0661 & 0.9247 & 0.0278 & 0.8869 & 0.7291 & 0.9607 & 0.3151 \\
DALE-CT-1S-v2    & 0.7854 & 0.0786 & 0.8733 & 0.0129 & \textbf{0.9271} & 0.8200 & 0.8704 & 0.1305 \\
DALE-CT-2S       & \textbf{0.8202} & \textbf{0.1160} & 0.9120 & \textbf{0.0438} & 0.9193 & \textbf{0.8554} & 0.9632 & \textbf{0.6118} \\
\bottomrule
\end{tabular}
\vspace{1ex}

\raggedright
\footnotesize \textit{Note: All metrics are macro-averaged over valid classes (minority classes with $< 5$ positive evaluation samples excluded); the best model per task/probe combination is bolded.}
\end{table*}

We first verify that the auxiliary supervision learns what it is trained on, probing the frozen backbones' dense 2D features directly: on 414 held-out, ReX-annotated CT-RATE validation scans (deterministic 70/10/20 split), slice-level (\texttt{[CLS]}) probes predict the 14 ReX abnormality classes or the 118 TS coverage classes, and patch-level probes predict the corresponding masks pooled to the ViT patch grid; each probe is a single linear layer trained with the BCE loss matching the pretraining auxiliary head (full settings in Appendix~\ref{app:probes}).

Table~\ref{tab:aux_probe_results_wide} reports the results. At the patch level, dual-source supervision more than doubles anatomical probing precision: DALE-CT-2S reaches a TS-Patch AUPRC of 0.6118, versus 0.2938 for the unsupervised DALE-CT-0 and 0.4337 for Finetuned DINOv2---the margin over a strong natural-image-initialized backbone indicates the gain comes from dense CT-specific supervision rather than backbone capacity. DALE-CT-2S likewise leads every AUPRC column, including patch-level localization of the sparse ReX findings. The single-source model distributes this information differently: DALE-CT-1S-v2 concentrates anatomical evidence in the \texttt{[CLS]} token (the best TS-CLS AUROC, 0.9271) while its patch tokens are the least linearly decodable of any model (TS-Patch AUPRC 0.1305)---adding the ReX objective in 2S restores dense patch structure without sacrificing slice-level strength. Scale provides a smaller, supervision-free version of the same effect, with DALE-CT-0-L improving over DALE-CT-0 on every patch-level probe. The DINOv2 baseline inverts this pattern. Its \texttt{[CLS]} probes are the weakest of any model (0.6636 ReX-CLS and 0.7683 TS-CLS AUROC), yet its patch tokens are the strongest of any supervision-free arm, beating both DALE-CT-0 and DALE-CT-0-L on all four patch probes and taking the best patch AUROC on both tasks (0.9473 ReX, 0.9644 TS). The objectives account for the split: iBOT's masked-token loss supervises patch tokens directly, whereas the LeJEPA invariance and SIGReg terms act on pooled projections, so dense structure emerges only indirectly unless the auxiliary head supplies it. Only DALE-CT-2S, which does supply it, overtakes DINOv2 at the patch level, and only on AUPRC.

\subsection{In-Domain Efficacy on CT-RATE}
\label{sec:3d_probe}
\label{sec:error_bars}
To compare the frozen backbones on equal footing, every model---including the public 3D baselines---is probed by us under a single linear-probing MIL protocol adapted from \cite{colipri}: we replace each encoder's native aggregation with a trained head, search a grid of learning rates and pooling schemes (average, max, and learned-attention), feed the pooled vector to a linear layer over the 18 CT-RATE abnormalities, select the best probe by validation AUPRC, and fit per-class F1-maximizing decision thresholds on the validation set. All models are evaluated on a common CT-RATE partition (4,942 train / 992 test scans), and we report 95\% confidence intervals from a 2,000-resample test-set bootstrap. Where intervals overlap, we describe the corresponding orderings as point estimates. Full grid, optimizer, token-aggregation, and bootstrap details are in Appendix~\ref{app:probes}; a side-by-side summary of every evaluated model is in Appendix~\ref{app:models}.

\begin{table*}[t]
\centering
\caption{In-Domain Test Metrics on CT-RATE (Macro Averages, $n_{\text{test}}=992$; 95\% bootstrap CIs in gray).}
\label{tab:ct_rate_results}
\footnotesize
\setlength{\tabcolsep}{4pt}
\begin{tabular}{lcccc}
\toprule
\rowcolor{ukblue!10}\textbf{Model} & \textbf{AUPRC} & \textbf{AUROC} & \textbf{Macro F1} & \textbf{BA} \\
\midrule
CT-CLIP \cite{ct-rate} & \makecell{0.2564 \\ {\color{gray}\scriptsize $[0.2451, 0.2803]$}} & \makecell{0.6040 \\ {\color{gray}\scriptsize $[0.5861, 0.6224]$}} & \makecell{0.2991 \\ {\color{gray}\scriptsize $[0.2822, 0.3152]$}} & \makecell{0.5394 \\ {\color{gray}\scriptsize $[0.5306, 0.5484]$}} \\
\addlinespace
CT-FM \cite{ct-fm} & \makecell{0.4211 \\ {\color{gray}\scriptsize $[0.4062, 0.4475]$}} & \makecell{0.7670 \\ {\color{gray}\scriptsize $[0.7555, 0.7779]$}} & \makecell{0.4423 \\ {\color{gray}\scriptsize $[0.4235, 0.4595]$}} & \makecell{0.6605 \\ {\color{gray}\scriptsize $[0.6489, 0.6723]$}} \\
\addlinespace
Merlin \cite{merlin} & \makecell{0.4631 \\ {\color{gray}\scriptsize $[0.4473, 0.4900]$}} & \makecell{0.7810 \\ {\color{gray}\scriptsize $[0.7695, 0.7915]$}} & \makecell{0.4660 \\ {\color{gray}\scriptsize $[0.4468, 0.4843]$}} & \makecell{0.6775 \\ {\color{gray}\scriptsize $[0.6665, 0.6891]$}} \\
\addlinespace
COLIPRI-CRM \cite{colipri} & \makecell{\textbf{0.5799} \\ {\color{gray}\scriptsize $[0.5623, 0.6030]$}} & \makecell{\textbf{0.8480} \\ {\color{gray}\scriptsize $[0.8395, 0.8566]$}} & \makecell{\textbf{0.5684} \\ {\color{gray}\scriptsize $[0.5509, 0.5844]$}} & \makecell{\textbf{0.7456} \\ {\color{gray}\scriptsize $[0.7348, 0.7567]$}} \\
\midrule
Finetuned DINOv2 & \makecell{0.4902 \\ {\color{gray}\scriptsize $[0.4726, 0.5164]$}} & \makecell{0.7953 \\ {\color{gray}\scriptsize $[0.7852, 0.8056]$}} & \makecell{0.4631 \\ {\color{gray}\scriptsize $[0.4423, 0.4810]$}} & \makecell{0.6736 \\ {\color{gray}\scriptsize $[0.6623, 0.6845]$}} \\
\addlinespace
\rowcolor{ukblue!5}DALE-CT-0 & \makecell{0.5112 \\ {\color{gray}\scriptsize $[0.4930, 0.5375]$}} & \makecell{0.8057 \\ {\color{gray}\scriptsize $[0.7952, 0.8159]$}} & \makecell{0.5011 \\ {\color{gray}\scriptsize $[0.4822, 0.5188]$}} & \makecell{0.7067 \\ {\color{gray}\scriptsize $[0.6954, 0.7182]$}} \\
\addlinespace
\rowcolor{ukblue!5}DALE-CT-0-L & \makecell{0.5206 \\ {\color{gray}\scriptsize $[0.5016, 0.5464]$}} & \makecell{0.8156 \\ {\color{gray}\scriptsize $[0.8047, 0.8254]$}} & \makecell{0.5054 \\ {\color{gray}\scriptsize $[0.4852, 0.5237]$}} & \makecell{0.7019 \\ {\color{gray}\scriptsize $[0.6901, 0.7135]$}} \\
\addlinespace
\rowcolor{ukblue!5}DALE-CT-1S-v2 & \makecell{0.5074 \\ {\color{gray}\scriptsize $[0.4898, 0.5323]$}} & \makecell{0.8098 \\ {\color{gray}\scriptsize $[0.8002, 0.8189]$}} & \makecell{0.4969 \\ {\color{gray}\scriptsize $[0.4780, 0.5147]$}} & \makecell{0.6996 \\ {\color{gray}\scriptsize $[0.6891, 0.7099]$}} \\
\addlinespace
\rowcolor{ukblue!5}DALE-CT-2S & \makecell{\textbf{0.5312} \\ {\color{gray}\scriptsize $[0.5134, 0.5556]$}} & \makecell{\textbf{0.8247} \\ {\color{gray}\scriptsize $[0.8152, 0.8333]$}} & \makecell{\textbf{0.5064} \\ {\color{gray}\scriptsize $[0.4871, 0.5239]$}} & \makecell{\textbf{0.7085} \\ {\color{gray}\scriptsize $[0.6969, 0.7195]$}} \\
\bottomrule
\end{tabular}
\vspace{1ex}

\raggedright
\footnotesize \textit{Note: All models are probed by us under one protocol on a common CT-RATE partition (Section~\ref{sec:error_bars}); each value is the single best probe (val-selected configuration) with a 2{,}000-resample bootstrap 95\% CI. Bold marks the best 3D baseline (top block) and the best 2D model (bottom block); BA is balanced accuracy.}
\end{table*}

\begin{figure*}[tp]
    \centering
    \includegraphics[width=\textwidth]{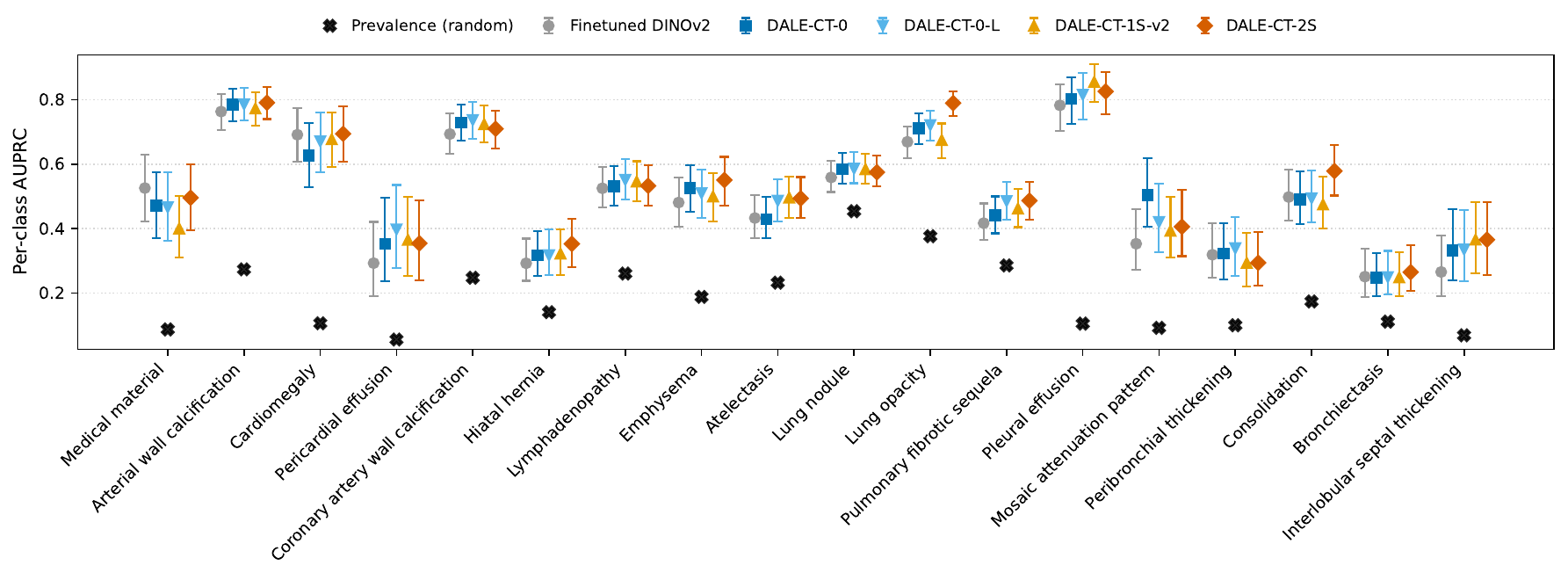}
    \caption{Per-class AUPRC on the CT-RATE test set for the five 2D backbones (5-seed mean with 95\% bootstrap confidence intervals on the mean). Black $\times$ markers denote per-class prevalence, the random-ranker baseline for AUPRC. Classes are shown in canonical order.}
    \label{fig:per_class_ctrate}
\end{figure*}

The performance of our evaluated models, alongside state-of-the-art public baselines, is presented in Table~\ref{tab:ct_rate_results}. Our results demonstrate that 2D slice-based architectures are competitive with state-of-the-art 3D VLMs on in-domain tasks. The top-performing 3D baseline, COLIPRI-CRM \cite{colipri}, establishes the upper bound with an area under the receiver operating characteristic curve (AUROC) of 0.8480 and an AUPRC of 0.5799. Our DALE-CT-2S configuration is the strongest 2D model, reaching an AUROC of 0.8247 and an AUPRC of 0.5312---within 0.024 AUROC of COLIPRI-CRM despite operating entirely without textual supervision or global volume metadata. The scaled DALE-CT-0-L appears in the table for completeness and is discussed in Section~\ref{sec:scale}. This places our 2D framework ahead of several established 3D models, including CT-CLIP \cite{ct-rate} (0.6040 AUROC), CT-FM \cite{ct-fm} (0.7670 AUROC), and Merlin \cite{merlin} (0.7810 AUROC), despite our pipeline operating entirely without global metadata or radiological reports during pretraining. Fig.~\ref{fig:per_class_ctrate} disaggregates these macro AUPRC averages to the per-class level for the five 2D backbones, with per-class prevalence shown as the random-ranker baseline; per-class AUROC for every evaluated model, on both datasets, is tabulated in Appendix~\ref{app:perclass}. At inference time these results are robust to the preprocessing pipeline: across seven crop, resolution, and tiling strategies the AUPRC spread is only 0.0235, with body-cropped $256\times256$ inference optimal (Appendix~\ref{app:preproc}).

A comparison of the 2D models shows that strong CT representations do not require natural-image pretraining. Continually pretraining a natural-image DINOv2 baseline on the CT-RATE cohort yields competitive representations (0.7953 AUROC), but a LeJEPA encoder trained from scratch on the same cohort does better. DALE-CT-0 achieves an AUROC of 0.8057 using no slice-level or global labels, completely bypassing the need for pretrained weights.

The addition of the dense auxiliary supervision mechanisms yields a monotonic improvement in AUROC point estimates. Introducing single-source anatomical guidance (DALE-CT-1S-v2) raises the in-domain AUROC point estimate from 0.8057 (DALE-CT-0) to 0.8098; the structural awareness this provides also serves specific generative downstream tasks exceptionally well \cite{bumgardner}. Pairing this anatomical guidance with pathology-specific ReX targets in our dual-source model (DALE-CT-2S) yields the most highly discriminative representations, pushing the in-domain AUROC to 0.8247 and our highest AUPRC of 0.5312. Because DALE-CT-1S-v2 shares the 2S architecture and dense TotalSegmentator objective with ReX held inert, the 0.0149 AUROC gap between them is attributable to the ReX abnormality supervision, although the two intervals overlap; the pattern is consistent with ReX helping in-domain probing without a commensurate gain on out-of-distribution transfer (Section~\ref{sec:external_gen}).

\begin{table*}[tp]
\centering
\caption{External Generalization Metrics on RAD-ChestCT (Macro Averages, $n_{\text{test}}=360$; 95\% bootstrap CIs on the seed mean).}
\label{tab:rad_chestct_results}
\scriptsize
\setlength{\tabcolsep}{2pt}
\begin{tabular}{lcccccccc}
\toprule
 & \multicolumn{4}{c}{\textbf{Frozen}} & \multicolumn{4}{c}{\textbf{Retrained}} \\
\cmidrule(lr){2-5} \cmidrule(lr){6-9}
\rowcolor{ukblue!10}\textbf{Model} & \textbf{AUPRC} & \textbf{AUROC} & \textbf{Macro F1} & \textbf{BA} & \textbf{AUPRC} & \textbf{AUROC} & \textbf{Macro F1} & \textbf{BA} \\
\midrule
CT-CLIP \cite{ct-rate} & \makecell{0.2861 \\ {\color{gray}\scriptsize $[0.2745, 0.3164]$}} & \makecell{0.5116 \\ {\color{gray}\scriptsize $[0.4860, 0.5369]$}} & \makecell{0.3676 \\ {\color{gray}\scriptsize $[0.3503, 0.3845]$}} & \makecell{0.4950 \\ {\color{gray}\scriptsize $[0.4848, 0.5048]$}} & \makecell{0.3339 \\ {\color{gray}\scriptsize $[0.3205, 0.3647]$}} & \makecell{0.5795 \\ {\color{gray}\scriptsize $[0.5591, 0.6013]$}} & \makecell{0.3899 \\ {\color{gray}\scriptsize $[0.3703, 0.4080]$}} & \makecell{0.5339 \\ {\color{gray}\scriptsize $[0.5186, 0.5490]$}} \\
\addlinespace
CT-FM \cite{ct-fm} & \makecell{0.3282 \\ {\color{gray}\scriptsize $[0.3141, 0.3577]$}} & \makecell{0.5721 \\ {\color{gray}\scriptsize $[0.5501, 0.5939]$}} & \makecell{0.3380 \\ {\color{gray}\scriptsize $[0.3213, 0.3546]$}} & \makecell{0.5090 \\ {\color{gray}\scriptsize $[0.5014, 0.5168]$}} & \makecell{0.4994 \\ {\color{gray}\scriptsize $[0.4766, 0.5348]$}} & \makecell{0.7273 \\ {\color{gray}\scriptsize $[0.7073, 0.7466]$}} & \makecell{0.4838 \\ {\color{gray}\scriptsize $[0.4537, 0.5079]$}} & \makecell{0.6277 \\ {\color{gray}\scriptsize $[0.6094, 0.6453]$}} \\
\addlinespace
Merlin \cite{merlin} & \makecell{0.4191 \\ {\color{gray}\scriptsize $[0.4003, 0.4514]$}} & \makecell{0.6566 \\ {\color{gray}\scriptsize $[0.6366, 0.6762]$}} & \makecell{0.4108 \\ {\color{gray}\scriptsize $[0.3868, 0.4300]$}} & \makecell{0.5934 \\ {\color{gray}\scriptsize $[0.5770, 0.6083]$}} & \makecell{0.4986 \\ {\color{gray}\scriptsize $[0.4760, 0.5342]$}} & \makecell{0.7177 \\ {\color{gray}\scriptsize $[0.6976, 0.7377]$}} & \makecell{0.4855 \\ {\color{gray}\scriptsize $[0.4543, 0.5110]$}} & \makecell{0.6350 \\ {\color{gray}\scriptsize $[0.6148, 0.6537]$}} \\
\addlinespace
COLIPRI-CRM \cite{colipri} & \makecell{\textbf{0.4899} \\ {\color{gray}\scriptsize $[0.4712, 0.5213]$}} & \makecell{\textbf{0.7469} \\ {\color{gray}\scriptsize $[0.7295, 0.7635]$}} & \makecell{\textbf{0.4818} \\ {\color{gray}\scriptsize $[0.4587, 0.5020]$}} & \makecell{\textbf{0.6553} \\ {\color{gray}\scriptsize $[0.6383, 0.6717]$}} & \makecell{\textbf{0.5806} \\ {\color{gray}\scriptsize $[0.5596, 0.6153]$}} & \makecell{\textbf{0.7982} \\ {\color{gray}\scriptsize $[0.7832, 0.8136]$}} & \makecell{\textbf{0.5544} \\ {\color{gray}\scriptsize $[0.5259, 0.5781]$}} & \makecell{\textbf{0.6893} \\ {\color{gray}\scriptsize $[0.6722, 0.7067]$}} \\
\addlinespace
\midrule
Finetuned DINOv2 & \makecell{0.3743 \\ {\color{gray}\scriptsize $[0.3564, 0.4107]$}} & \makecell{0.6252 \\ {\color{gray}\scriptsize $[0.6034, 0.6463]$}} & \makecell{0.3455 \\ {\color{gray}\scriptsize $[0.3193, 0.3688]$}} & \makecell{0.5459 \\ {\color{gray}\scriptsize $[0.5305, 0.5615]$}} & \makecell{0.5538 \\ {\color{gray}\scriptsize $[0.5271, 0.5925]$}} & \makecell{0.7550 \\ {\color{gray}\scriptsize $[0.7367, 0.7729]$}} & \makecell{0.5276 \\ {\color{gray}\scriptsize $[0.4983, 0.5513]$}} & \makecell{0.6596 \\ {\color{gray}\scriptsize $[0.6421, 0.6761]$}} \\
\addlinespace
\rowcolor{ukblue!5}DALE-CT-0 & \makecell{0.3565 \\ {\color{gray}\scriptsize $[0.3396, 0.3921]$}} & \makecell{0.5946 \\ {\color{gray}\scriptsize $[0.5706, 0.6177]$}} & \makecell{0.2664 \\ {\color{gray}\scriptsize $[0.2423, 0.2891]$}} & \makecell{0.5305 \\ {\color{gray}\scriptsize $[0.5199, 0.5410]$}} & \makecell{0.5414 \\ {\color{gray}\scriptsize $[0.5143, 0.5785]$}} & \makecell{0.7477 \\ {\color{gray}\scriptsize $[0.7290, 0.7670]$}} & \makecell{0.5169 \\ {\color{gray}\scriptsize $[0.4861, 0.5434]$}} & \makecell{0.6492 \\ {\color{gray}\scriptsize $[0.6318, 0.6672]$}} \\
\addlinespace
\rowcolor{ukblue!5}DALE-CT-0-L & \makecell{\underline{0.3972} \\ {\color{gray}\scriptsize $[0.3758, 0.4344]$}} & \makecell{0.6281 \\ {\color{gray}\scriptsize $[0.6071, 0.6472]$}} & \makecell{\underline{0.3519} \\ {\color{gray}\scriptsize $[0.3291, 0.3719]$}} & \makecell{\underline{0.5487} \\ {\color{gray}\scriptsize $[0.5369, 0.5602]$}} & \makecell{\underline{0.5551} \\ {\color{gray}\scriptsize $[0.5259, 0.5936]$}} & \makecell{\underline{0.7572} \\ {\color{gray}\scriptsize $[0.7386, 0.7766]$}} & \makecell{\underline{0.5281} \\ {\color{gray}\scriptsize $[0.4968, 0.5556]$}} & \makecell{\underline{0.6588} \\ {\color{gray}\scriptsize $[0.6400, 0.6766]$}} \\
\addlinespace
\rowcolor{ukblue!5}DALE-CT-1S-v2 & \makecell{0.3886 \\ {\color{gray}\scriptsize $[0.3696, 0.4228]$}} & \makecell{\underline{0.6284} \\ {\color{gray}\scriptsize $[0.6083, 0.6474]$}} & \makecell{0.3489 \\ {\color{gray}\scriptsize $[0.3225, 0.3717]$}} & \makecell{0.5440 \\ {\color{gray}\scriptsize $[0.5320, 0.5559]$}} & \makecell{0.5207 \\ {\color{gray}\scriptsize $[0.4939, 0.5604]$}} & \makecell{0.7334 \\ {\color{gray}\scriptsize $[0.7131, 0.7528]$}} & \makecell{0.5044 \\ {\color{gray}\scriptsize $[0.4734, 0.5298]$}} & \makecell{0.6450 \\ {\color{gray}\scriptsize $[0.6263, 0.6631]$}} \\
\addlinespace
\rowcolor{ukblue!5}DALE-CT-2S & \makecell{0.3820 \\ {\color{gray}\scriptsize $[0.3598, 0.4215]$}} & \makecell{0.6252 \\ {\color{gray}\scriptsize $[0.6044, 0.6450]$}} & \makecell{0.1893 \\ {\color{gray}\scriptsize $[0.1716, 0.2053]$}} & \makecell{0.5322 \\ {\color{gray}\scriptsize $[0.5239, 0.5402]$}} & \makecell{0.5377 \\ {\color{gray}\scriptsize $[0.5135, 0.5742]$}} & \makecell{0.7389 \\ {\color{gray}\scriptsize $[0.7193, 0.7582]$}} & \makecell{0.5222 \\ {\color{gray}\scriptsize $[0.4945, 0.5458]$}} & \makecell{0.6480 \\ {\color{gray}\scriptsize $[0.6317, 0.6640]$}} \\
\bottomrule
\end{tabular}

\vspace{1ex}
\raggedright
\footnotesize \textit{Note: All models are probed by us on the common RAD-ChestCT test set (Section~\ref{sec:error_bars}); each value is the mean over the available probe seeds (1--5 per model) with a 95\% bootstrap CI on that mean in gray. Bold marks the best overall model per column; \underline{underline} marks the best DALE-CT model per column. ``Frozen'' applies the CT-RATE-trained classifier directly to RAD-ChestCT; ``Retrained'' keeps the encoder frozen and trains a new classifier on the RAD-ChestCT train split. Frozen-arm Macro F1 is threshold-sensitive on RAD-ChestCT's small validation set and should be read alongside AUPRC/AUROC.}
\end{table*}

\subsection{External Generalization and Classifier Alignment}
\label{sec:external_gen}

Testing on the external RAD-ChestCT dataset initially showed that our 2D models struggled when applied directly. As shown in Table~\ref{tab:rad_chestct_results}, the 2D models experienced a larger performance drop than the 3D baselines. For example, our best in-domain model (DALE-CT-2S) fell to a frozen-arm AUROC of 0.6252, whereas the 3D COLIPRI-CRM \cite{colipri} maintained a 0.7469 AUROC. Under the frozen protocol DALE-CT-1S-v2 (0.6284) and the scaled DALE-CT-0-L (0.6281) post the highest 2D point estimates, with DALE-CT-2S and Finetuned DINOv2 (both 0.6252) inside overlapping intervals.

We initially suspected our 2D architecture lacked broader 3D context. To test whether the models' underlying features were flawed or if the classifier simply struggled with different hospital scanner settings, we ran a second experiment: we kept the models' core features fixed and retrained the classifier on the new RAD-ChestCT dataset.

This retraining confirmed the features were adaptable rather than fundamentally limited: the 2D models recovered substantially. On the retrained arm, however, the 3D COLIPRI-CRM \cite{colipri} remains the strongest model at 0.7982 AUROC and 0.5806 AUPRC, ahead of every 2D backbone. Among the 2D models, the scaled DALE-CT-0-L posts the best point estimates (0.7572 AUROC, 0.5551 AUPRC), with Finetuned DINOv2 (0.7550/0.5538) inside overlapping intervals and the CT-RATE-trained DALE family close behind (DALE-CT-0 0.7477, DALE-CT-2S 0.7389, DALE-CT-1S-v2 0.7334). The retrained 2D models nonetheless outperform the weaker 3D baselines---CT-CLIP (0.5795), CT-FM (0.7273), and Merlin (0.7177)---confirming that the 2D representations transfer competitively despite the absence of textual supervision. Finetuned DINOv2's parity here is notable alongside Section~\ref{sec:aux_verify}: it transfers as well as our best scaled model while retaining markedly stronger dense features, making it the better frozen backbone for patch-level downstream tasks. Per-class results for both arms are tabulated in Appendix~\ref{app:perclass}.

Finally, this experiment revealed a trade-off with our training methods. While our most strictly guided model (DALE-CT-2S) performed best on the original dataset, on the retrained external arm its point estimate trails models trained with lighter or no auxiliary targets---most notably the scaled, no-auxiliary DALE-CT-0-L at 0.7572, alongside unguided DALE-CT-0 at 0.7477 and Finetuned DINOv2 at 0.7550. This suggests that highly specific, guided training targets can cause the model to specialize toward the original dataset's distribution. On the frozen arm, no 2D backbone transfers well: all sit between 0.59 and 0.63 AUROC, well below COLIPRI-CRM's 0.7469, with the differences among them inside overlapping intervals. As the retraining experiment shows, this deficit reflects classifier misalignment across cohorts rather than the quality of the underlying features. Finetuned DINOv2's retrained point estimate also slightly exceeds the unguided DALE-CT-0's; the intervals overlap, but such an edge would be consistent with the robust foundational feature space DINOv2 inherited from its pretrained weights \cite{dinov2-registers}. Combined with the practical constraint that ReX labels exist only on CT-RATE, this trade-off made the supervision-free configuration the natural candidate for scaling (Section~\ref{sec:scale}).

\subsection{Scaling the Supervision-Free Configuration}
\label{sec:scale}

The preceding results shaped our scaled model: dense supervision buys in-domain accuracy but its gains do not carry over externally, and the ReX-guided variants are bounded by label availability, since the multi-source pool carries ReX labels only on its CT-RATE subset. We therefore scaled the no-auxiliary configuration: DALE-CT-0-L pretrains the DALE-CT-0 recipe (Section~\ref{sec:augmentation}) on the $\sim$287k-scan multi-source corpus of Section~\ref{sec:large_corpus}, to our knowledge the largest chest-CT pretraining corpus reported to date.

On CT-RATE linear probing (Table~\ref{tab:ct_rate_results}), scaling raises the no-auxiliary model's point estimates from 0.8057 to 0.8156 AUROC and 0.5112 to 0.5206 AUPRC (within overlapping intervals), matching the supervised DALE-CT-1S-v2 (0.8098 AUROC) within error despite using no auxiliary labels. On the external RAD-ChestCT test set it posts the best 2D retrained point estimates (Table~\ref{tab:rad_chestct_results}). The dense probes of Table~\ref{tab:aux_probe_results_wide} show that scaling redistributes rather than uniformly improves this information: slice-level probing of the auxiliary targets degrades slightly (TS-CLS AUROC 0.9151 to 0.8869, ReX-CLS 0.7457 to 0.7241) while patch-level probing improves on both tasks (TS-Patch AUPRC 0.2938 to 0.3151, ReX-Patch 0.0177 to 0.0278).

Crucially, scale preserves the anatomical world model rather than diluting it. Frozen DALE-CT-0-L representations linearly decode slice position with $R^2{=}0.973$ ($\rho{=}0.997$, MAE $10.8$ slices), on par with the ViT-Base depth-aware ($R^2{=}0.971$) and pure-2D ($R^2{=}0.974$) ablations (Table~\ref{tab:world_model}), with bootstrap CIs that overlap. Slice ordering is likewise retained (PCA-1D $|\tau|{=}0.559$, the cleanest of the three; greedy-NN $|\tau|{=}0.428$). Organ identity beyond $z$-position is also retained: under the residualized clustering probe of Appendix~\ref{app:probes}, DALE-CT-0-L reaches NMI $0.725$ / ARI $0.606$, far above the pure-2D ablation ($0.458$/$0.258$) and within the depth-aware regime of the ViT-Base ablation ($0.755$/$0.662$) and supervised DALE-CT-2S ($0.765$/$0.673$). Its bootstrap CIs fall just below those two depth-aware backbones: the coarse $z$-coordinate is fully preserved at $11{\times}$ scale, while fine-grained organ-identity granularity shows a modest decrement rather than a collapse. (We restrict these comparisons to probes that are not confounded by embedding effective rank; see Appendix~\ref{app:probes}.)

We are careful not to over-claim the $25\text{k}{\to}287\text{k}$ contrast as a controlled scaling law. DALE-CT-0-L differs from the released DALE-CT-0 along several axes simultaneously (Table~\ref{tab:augmentation_configs}, Appendix~\ref{app:training}): corpus (multi-source vs.\ CT-RATE), slab instantiation (3-slice native-$z_1$ vs.\ 12~mm physical), ReX guidance (absent vs.\ $0.8$), normalization statistics (fit to the full pool vs.\ CT-RATE, via the identical 0.5/99.5-percentile procedure), and schedule (3 epochs / 191k iterations / batch 384 on 16 H100s vs.\ 50k iterations / batch 512 on 8 H200s). The contrast therefore reflects the full configuration of a broader-corpus instantiation, not corpus size in isolation. This is consistent with the field, where scaling behavior for CT foundation models remains largely unexplored; prior work reports only data-quantity curves (Merlin~\cite{merlin}; CT-CLIP~\cite{ct-rate}; FlexiCT~\cite{flexict}).

\section{Conclusion}
This work shows that a 2D Vision Transformer, trained from scratch with the heuristics-free LeJEPA objective and depth-aware slab sampling, learns an anatomical world model of chest CT---volumetric structure that an otherwise-identical encoder trained on slices in isolation never develops, emerging with no 3D or positional supervision anywhere in training. The same slice-based design sidesteps the aggressive resizing and interpolation that variable-length clinical scans force on fixed-geometry 3D models. Depth-aware training adds this structure without giving anything up. Classification accuracy is fully preserved and holds even on sparsely sampled scans, a robustness that matters where clinical slice spacing varies widely, and the smoother geometry produces the cleanest organ-coverage curves we measured, at some cost to boundary sharpness. Because our two frozen probes did not translate that geometry into higher downstream accuracy, we present the world model as a property of the representation and establish performance separately in a head-to-head frozen-probe benchmark. There, dense auxiliary supervision makes DALE-CT-2S the strongest 2D model in-domain, within 0.024 AUROC of COLIPRI-CRM without any text supervision, but that advantage does not survive the move to an external cohort, a trade-off that led us to scale the supervision-free configuration rather than the supervised one. DALE-CT-0-L retains the anatomical world model at $11\times$ the pretraining corpus and posts the best 2D external-transfer point estimates. We release it as the recommended general-purpose backbone, alongside the full model family, training code, and evaluation pipeline; where dense, patch-level features drive the task, the continually pretrained DINOv2 arm we release with it is the stronger choice. Together these give the medical imaging community a flexible, efficient, and structurally grounded foundation for diagnostic and generative pipelines.

\section*{Code and Model Availability}

To support open science and reproducibility, all training code, evaluation pipelines, and pretrained model weights discussed in this work are publicly available. The core training framework and LeJEPA implementation are available in the \href{https://github.com/Kentucky-Open-Science/DALE-CT}{\textbf{DALE-CT repository}}. The head-to-head evaluation harness is released separately as the \href{https://github.com/Kentucky-Open-Science/chest-ct-foundation-model-benchmark}{\textbf{chest-ct-foundation-model-benchmark}} repository, which distributes the unified backbone loader that normalizes every evaluated encoder behind one interface, the patient-disjoint splits, the frozen-probe protocol, and every benchmark result reported here. The benchmark is maintained as a living resource and continues to add encoders beyond those evaluated in this paper; the numbers reported here correspond to a fixed snapshot of its results.

All pretrained weights are hosted on the Hugging Face Hub under CC-BY-NC-SA-4.0 and load directly via \texttt{timm}, e.g.,
\begin{center}
\footnotesize\texttt{timm.create\_model(\char`\"{}hf-hub:Kentucky-Open-\linebreak Science/DALE-CT-0-L\char`\"{}, pretrained=True)}
\end{center}
The released models are:
\begin{itemize}
    \item \href{https://huggingface.co/Kentucky-Open-Science/DALE-CT-0-L}{\textbf{DALE-CT-0-L}} --- the recommended general-purpose backbone (best external-transfer point estimates; supervision-free at scale)
    \item \href{https://huggingface.co/Kentucky-Open-Science/DALE-CT-2S}{\textbf{DALE-CT-2S}} --- the strongest in-domain (CT-RATE) model
    \item \href{https://huggingface.co/Kentucky-Open-Science/DALE-CT-1S-v2}{\textbf{DALE-CT-1S-v2}}
    \item \href{https://huggingface.co/Kentucky-Open-Science/DALE-CT-0}{\textbf{DALE-CT-0}}
    \item \href{https://huggingface.co/Kentucky-Open-Science/Finetuned-DINOv2-Chest-CT}{\textbf{Finetuned DINOv2}} --- the strongest dense (patch-level) features; ties DALE-CT-0-L on external transfer
\end{itemize}
The earlier \href{https://huggingface.co/Kentucky-Open-Science/DALE-CT-1S}{DALE-CT-1S} release (patch-14, \texttt{[CLS]}-only anatomical supervision) remains available as the backbone used by Ker-VLJEPA-3B \cite{bumgardner}; the DALE-CT-1S-v2 repository holds the configuration reported in this paper.

\bibliographystyle{IEEEtran}
\bibliography{bib} 

\appendices

\section{SIGReg Formulation}
\label{app:sigreg}

SIGReg matches the empirical characteristic function (CF) of the projected embeddings to that of a standard Gaussian. Let $A = \{a_1,\dots,a_K\}$ be $K{=}1024$ random unit-norm directions in $\mathbb{R}^D$, resampled each step with seeds synchronized across devices. For each direction $a$, the scalar projections $\{a^\top h_{n,v}\}$ have empirical CF $\hat\varphi_a(t) = \frac{1}{N}\sum_i e^{\,i t\, a^\top h_i}$, where $N$ is the global minibatch size after all-reduce. SIGReg is the Gaussian-weighted $L_2$ distance between $\hat\varphi_a$ and the standard-Gaussian CF $\varphi(t)=e^{-t^2/2}$, integrated by the trapezoidal rule over 17 knots on $[-3,3]$ and averaged over directions,
\begin{equation}
\mathcal{L}_{\text{SIGReg}} = \frac{1}{K}\sum_{a\in A}\; N\!\int_{-3}^{3} \big|\hat\varphi_a(t) - e^{-t^2/2}\big|^2\, e^{-t^2/2}\, dt.
\end{equation}

\section{Training Configuration}
\label{app:training}

All DALE-CT variants use a ViT-Large, patch-16 backbone initialized from scratch and trained with Distributed Data Parallel in \texttt{bf16} mixed precision. DALE-CT-0 and DALE-CT-2S train for 50{,}000 iterations on 8$\times$H200 GPUs (global batch 512, 64 per GPU; 5{,}000-step warmup; peak learning rate $3\!\times\!10^{-4}$ decaying to $3\!\times\!10^{-5}$ under cosine annealing). DALE-CT-1S-v2 trains for 66{,}667 iterations on 8$\times$H100 GPUs (global batch 384, 48 per GPU; 6{,}667-step warmup; same schedule), preserving the 25.6M-image budget of DALE-CT-2S ($48\!\times\!8\!\times\!66{,}667 \approx 64\!\times\!8\!\times\!50{,}000$), so the CT-RATE supervision ladder (0 $\to$ 1S-v2 $\to$ 2S) is architecture-, patch-size-, and budget-controlled. DALE-CT-0-L trains for 3 epochs over the multi-source pool (191{,}357 iterations, global batch 384 on 16$\times$H100 GPUs) with the auxiliary head disabled.

Every crop of the multi-crop set (Section~\ref{sec:augmentation}) is resized to its target resolution (bilinear interpolation for the image, nearest-neighbor for the annotation masks), then passed through a random horizontal flip ($p{=}0.5$), a random gamma adjustment ($\gamma\sim\mathcal{U}(0.9,1.1)$, applied with $p{=}0.8$), an optional $3{\times}3$ Gaussian blur, and z-score normalization. Following the DINO convention, the two global views are blurred with probabilities $0.5$ and $0.1$ and the eight local views are never blurred. Table~\ref{tab:augmentation_configs} contrasts the two pretraining regimes.

\begin{table}[t]
\centering
\caption{Slab, guidance, and normalization contrast for the two DALE-CT pretraining regimes.}
\label{tab:augmentation_configs}
\resizebox{\columnwidth}{!}{%
\small
\begin{tabular}{lll}
\toprule
\rowcolor{ukblue!10} & \textbf{CT-RATE models} & \textbf{DALE-CT-0-L} \\
\rowcolor{ukblue!10} & (0 / 1S-v2 / 2S) & (full pool) \\
\midrule
Pretraining corpus & 25{,}692 CT-RATE & $\sim$287k multi-source \\
Slab instantiation & 12~mm physical & 3-slice native-$z_1$ \\
Local-crop guidance & ReX $p{=}0.8$ + TS & TS foreground \\
Norm clip [HU] & $[-997,\,888]$ & $[-940.8,\,923.1]$ \\
Norm $\mu,\sigma$ & $-142,\,361$ & $-25.03,\,246.87$ \\
\bottomrule
\end{tabular}}
\vspace{1ex}
\raggedright
\footnotesize \textit{Note: Crop geometry (two $256^2$ global and eight $144^2$ local crops), geometric transforms, and the SIGReg regularizer ($\lambda{=}0.02$) are identical across regimes; only the slab instantiation, label guidance, normalization statistics, and corpus differ. Normalization uses the same 0.5/99.5-percentile clip then standardization in both, fit to each corpus independently.}
\end{table}

\section{Attribution Study Protocols}
\label{app:attribution}

\textbf{The matched pair.} Pure-2D ViT-Base and depth-aware ViT-Base are the CT-RATE ablation backbones of Section~\ref{sec:world_model}, identical in architecture, initialization, corpus, and budget; the arms differ only in whether self-supervised views are sampled within a single slice or across a 12~mm slab.

\textbf{Organ-span localization.} A RidgeCV probe maps each frozen per-slice \texttt{[CLS]} embedding to 118-class TotalSegmentator fractional coverage, yielding for every organ a predicted coverage curve along the scan's slice axis. The probed organs are the 12 largest thoracic-window structures: heart, left lung, right lung, liver, spleen, stomach, left kidney, right kidney, esophagus, trachea, aorta, and sternum. \emph{Largest-run extraction}: a scan- and organ-specific threshold is set as a fraction of that curve's maximum (the fraction tuned per organ on the validation split); every slice above the threshold is marked, contiguous marked slices form runs, and the longest run is taken as the predicted span---so isolated single-slice excursions (flicker) do not fragment the prediction, and the number of supra-threshold runs is itself reported as the flicker measure. Metrics: span IoU against the TotalSegmentator-derived true range, boundary MAE in slices, and the run count; paired per-organ bootstrap deltas over the 86 test scans of the 414-scan ReX-annotated split (70/10/20). Both the probe and its ground truth derive from TotalSegmentator, so the span results speak to inference cost rather than accuracy against an independent reference. Table~\ref{tab:expf_per_organ} reports the per-organ results.

\begin{table}[htbp]
\centering
\caption{Per-organ span IoU (86 test scans). $\Delta$ = paired depth-aware $-$ pure-2D span-IoU delta with 95\% bootstrap CI; bold where the CI excludes zero.}
\label{tab:expf_per_organ}
\setlength{\tabcolsep}{3pt}
\resizebox{\columnwidth}{!}{%
\begin{tabular}{lccc}
\toprule
\rowcolor{ukblue!10}\textbf{Organ} & \textbf{Pure-2D} & \textbf{Depth-aware} & $\Delta$ \textbf{(DA$-$2D)} \\
\midrule
Heart & 0.904 & 0.887 & $-0.017$ {\scriptsize $[-0.036,+0.002]$} \\
Lung (L) & 0.791 & 0.757 & \textbf{$-0.033$ {\scriptsize $[-0.059,-0.011]$}} \\
Lung (R) & 0.877 & 0.829 & \textbf{$-0.047$ {\scriptsize $[-0.079,-0.023]$}} \\
Liver & 0.890 & 0.832 & \textbf{$-0.058$ {\scriptsize $[-0.093,-0.029]$}} \\
Spleen & 0.801 & 0.781 & $-0.020$ {\scriptsize $[-0.048,+0.008]$} \\
Stomach & 0.725 & 0.647 & \textbf{$-0.078$ {\scriptsize $[-0.117,-0.041]$}} \\
Kidney (L) & 0.629 & 0.545 & \textbf{$-0.084$ {\scriptsize $[-0.130,-0.039]$}} \\
Kidney (R) & 0.851 & 0.796 & \textbf{$-0.055$ {\scriptsize $[-0.078,-0.032]$}} \\
Esophagus & 0.862 & 0.817 & \textbf{$-0.045$ {\scriptsize $[-0.073,-0.016]$}} \\
Trachea & 0.754 & 0.694 & \textbf{$-0.060$ {\scriptsize $[-0.096,-0.022]$}} \\
Aorta & 0.704 & 0.678 & $-0.026$ {\scriptsize $[-0.058,+0.005]$} \\
Sternum & 0.792 & 0.737 & \textbf{$-0.054$ {\scriptsize $[-0.077,-0.032]$}} \\
\midrule
Macro & 0.798 & 0.750 & \\
\bottomrule
\end{tabular}}
\end{table}

\textbf{Stride robustness.} The evaluation loads the exact grid-search winners behind Table~\ref{tab:ablation_results}---for each arm, the learned-attention probe (selected by validation AUPRC from a grid of six learning rates and three pooling schemes) and its validation-fit per-class F1 thresholds, unmodified---and re-evaluates them on the same full CT-RATE validation split (3{,}002 scans), with each test bag subsampled to every $k$-th slice ($k\in\{1,2,4,8\}$, offset 0). Probes are never re-fit. The $k{=}1$ column reproduces both arms' original test metrics to four decimal places, validating the harness.

\section{Probing Protocol Details}
\label{app:probes}

\begin{figure*}[tp]
    \centering
    \includegraphics[width=\textwidth]{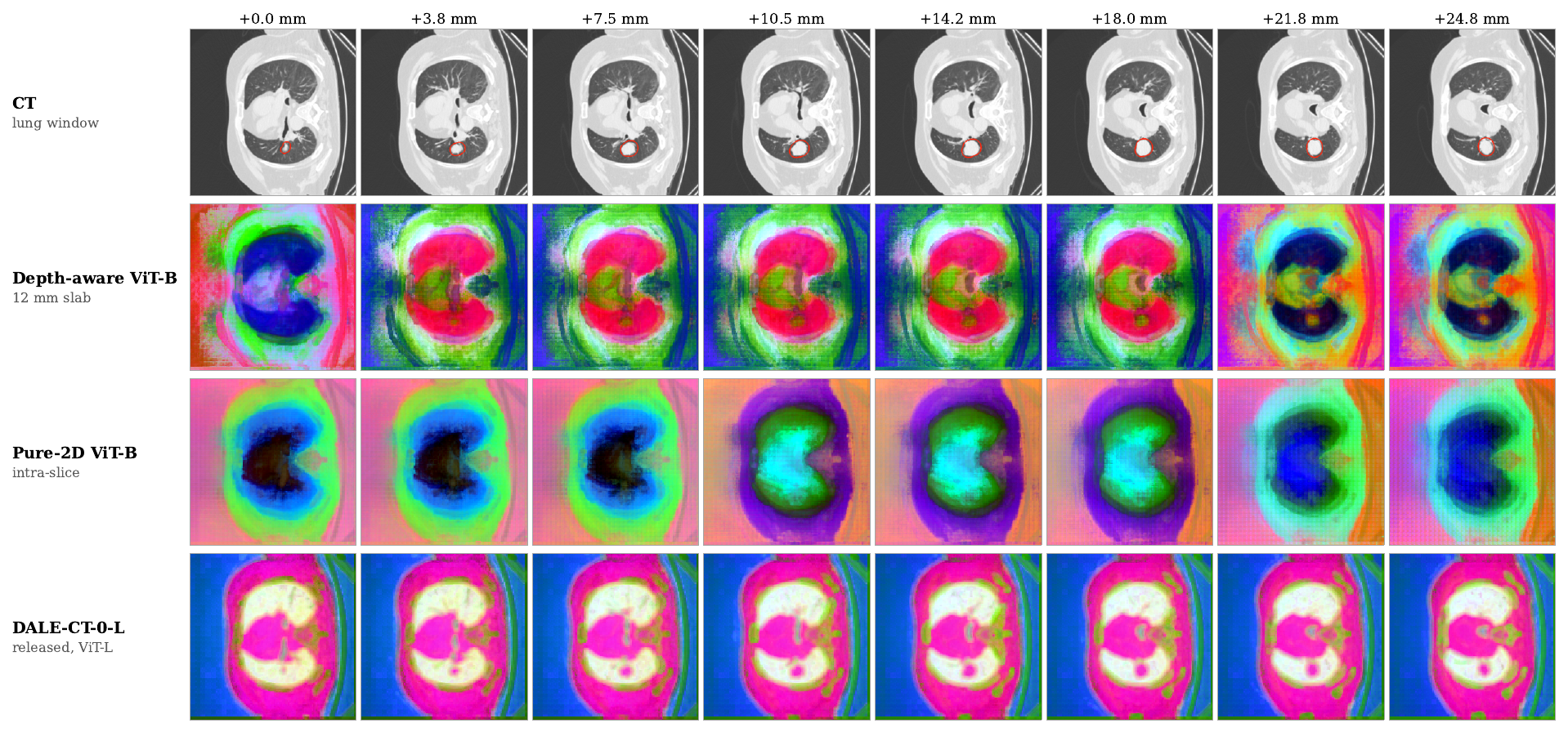}
    \caption{\textbf{Per-slice PCA control for Fig.~\ref{fig:z_continuity}.} Identical volume, slices, backbones and offset-ensembled lattice, but with the PCA basis fit independently on each slice instead of shared across them. Given its own basis, the pure-2D backbone resolves lung, mediastinum and chest wall as clearly as the others: its features are \emph{not} weak or uninformative, they are simply not registered to a common basis across $z$, so the components describing one slice do not describe the next. Colors are therefore not comparable between columns, which is why the shared basis of Fig.~\ref{fig:z_continuity} is the appropriate rendering for a continuity claim. DALE-CT-0-L is nearly unchanged between the two figures, indicating its per-slice bases already agree.}
    \label{fig:z_continuity_perslice}
\end{figure*}

\textbf{Organ identity beyond $z$ (residualized clustering).} The $z$-residualized organ-identity probe (Sections~\ref{sec:world_model} and~\ref{sec:scale}) is constructed as follows. On the 414-scan ReX-annotated split, a ridge regressor (regularization selected by cross-validation over $10^{-3}$--$10^{3}$) is fit on the training scans from per-slice \texttt{[CLS]} embeddings to normalized within-scan position $z\in[0,1]$; its weight vector defines the dominant linear position direction, and each test-scan embedding is residualized by projecting out this single direction. Within each test scan, two independent $k$-means clusterings ($k{=}5$ anatomical bands; 10 restarts, fixed seed) are then computed: one over the $z$-residualized \texttt{[CLS]} embeddings, and one over the binarized 118-dimensional TotalSegmentator organ-presence vector of each slice. Agreement between the latent and anatomical clusterings is scored per scan by NMI and ARI, averaged over test scans (scans with fewer than $k{+}2$ slices are skipped), with 2{,}000-resample bootstrap CIs. Because only one linear direction is removed, the probe controls for the trivial linear $z$ signal, not for all position information; nonlinear position correlates may remain. The ViT-Base ablation arms carry no auxiliary head, so neither saw organ labels during pretraining; DALE-CT-2S (compared in Section~\ref{sec:scale}) did, and is labeled as supervised there.

\textbf{Dense 2D probing (Section~\ref{sec:aux_verify} evaluation).} Slice-level (\texttt{[CLS]}) labels are a 14-class binary vector ($\mathbf{y}\in\{0,1\}^{14}$) for ReX and a 118-class soft fractional-coverage vector ($\mathbf{y}\in[0,1]^{118}$) for TotalSegmentator; patch-level labels are the segmentation masks max-pooled to the ViT patch grid ($G\times G$), giving $[14,G,G]$ binary and $[118,G,G]$ soft tensors. Slices are body-cropped and resized to $256\times256$ ($252\times252$ for the DINOv2 baseline, for patch divisibility), and we extract per-slice \texttt{[CLS]} tokens $[D,1024]$ and flattened patch tokens $[D,N_{\text{patches}},1024]$. Each probe is a single linear layer trained with multi-label BCE for 15{,}000 steps (batch 256, SGD with 0.95 momentum, cosine annealing), selecting the learning rate from $\{0.1,0.03,0.01,0.003\}$ by validation macro AUPRC.

\textbf{3D MIL probing (Section~\ref{sec:3d_probe}).} The grid search spans six learning rates ($\{0.3, 0.1, 0.03, 0.01, 0.003, 0.001\}$) and three pooling schemes (average, max, and learned-attention, a subset of the five in \cite{colipri}). The pooled vector feeds a linear layer over the 18 CT-RATE abnormalities, optimized for 15{,}000 steps with SGD (batch 16, momentum 0.95, weight decay 0, cosine annealing). The best probe is selected by validation AUPRC, followed by a per-class F1-maximizing decision threshold fit on the validation set.

\textbf{Token aggregation and uncertainty.} The 3D baselines contribute a flattened grid of volumetric tokens from a centered crop, and the 2D models contribute a per-slice \texttt{[CLS]} token bag; average, max, and learned-attention pooling reduce either to a single vector. Merlin is the exception: its image encoder does not expose a dense token grid, returning a single pooled volume-level embedding that we probe directly (the three pooling schemes coincide on a single token). Bootstrap resample indices are drawn deterministically from a per-task fixed seed and are shared across all models within a task, so the confidence intervals are directly comparable and paired differences between models remain available; the external arms use the common RAD-ChestCT test set ($n_{\text{test}}=360$).

\textbf{Effective rank and geometric-probe confounds.} Depth-aware training concentrates the embedding spectrum: DALE-CT-0-L has a participation ratio of $4.16$, versus $16.7$ for the pure-2D ViT-Base ablation and $5.79$ for the ViT-Base depth-aware ablation. Probes whose values depend on raw pairwise distances therefore conflate representational content with effective rank, and we do not interpret them as scale effects. The two probes reported in Section~\ref{sec:scale}---supervised $z$-position regression and organ-identity clustering after residualizing out $z$---are not rank-confounded, which is why the scale comparison is restricted to them.

\section{Characteristics of Evaluated Models}
\label{app:models}

Table~\ref{tab:model_comparison} summarizes the architecture, pretraining corpus, and pretraining objective of every model evaluated in this work.

\begin{table*}[tp]
\centering
\caption{Characteristics of Evaluated Volumetric Foundation Models and Baselines}
\label{tab:model_comparison}
\resizebox{\textwidth}{!}{%
\begin{tabular}{@{}lllp{4cm}p{6cm}@{}}
\toprule
\rowcolor{ukblue!10}\textbf{Model} & \textbf{Architecture} & \textbf{Training Data (Amount)} & \textbf{Training Data (Type)} & \textbf{Pretraining Technique} \\
\midrule
\multicolumn{5}{c}{\textbf{3D Foundation Models}} \\
\midrule
\textbf{CT-FM} \cite{ct-fm} & 3D SegResNet & 148,394 unique CT scans & Unannotated whole-body 3D CT scans & Task-agnostic, intra-sample 3D image-based contrastive self-supervised learning (SSL) \\
\addlinespace
\textbf{CT-CLIP} \cite{ct-rate} & 3D ViT \& CXR-Bert & CT-RATE cohort (25,692 unique scans) & Non-contrast 3D chest CTs paired with free-text radiology reports & 3D Contrastive Language-Image Pretraining (CLIP) to map volumes and text into a shared latent space \\
\addlinespace
\textbf{Merlin} \cite{merlin} & 3D ResNet-152 \& Longformer & 15,331 unique CT scans & Abdominal CTs paired with structured EHR diagnosis codes and radiology reports & Multitask vision-language pretraining (InfoNCE for reports, BCE for EHR diagnosis codes) \\
\addlinespace
\textbf{COLIPRI-CRM} \cite{colipri} & 3D Primus-M ViT \& CXR-BERT & $\sim$97,000 unique scans & Combination of paired 3D chest CTs/reports (CT-RATE) and unpaired chest CTs (NLST) & Alternating 3D CLIP loss, autoregressive Radiology Report Generation, and Masked Image Modeling (MIM) \\
\midrule
\multicolumn{5}{c}{\textbf{2D Slice-Based Volumetric Models (Ours)}} \\
\midrule
\textbf{Finetuned DINOv2} & 2D ViT-Large (Registers) & CT-RATE cohort (25,692 unique scans) & 2D CT slices grouped into 12 mm slabs (foreground isolated) & Continuous DINO-MX distillation + iBOT masked image modeling from natural image weights \\
\addlinespace
\textbf{DALE-CT-0} & 2D ViT-Large (patch16) & CT-RATE cohort (25,692 unique scans) & 2D CT slices dynamically sampled into multi-crop global/local views & From-scratch invariance SSL with Sketched Isotropic Gaussian Regularization (SIGReg) \\
\addlinespace
\textbf{DALE-CT-0-L} & 2D ViT-Large (patch16) & Multi-source chest-CT pool (287,302 training scans, 32 collections) & Multi-source 2D CT slices dynamically sampled into multi-crop views across a 3-slice axial slab (native-$z_1$; no ReX guidance) & From-scratch invariance SSL with SIGReg; no auxiliary supervision (full-pool scale-up of DALE-CT-0) \\
\addlinespace
\textbf{DALE-CT-1S-v2} & 2D ViT-Large (patch16) & CT-RATE cohort (25,692 unique scans) & 2D CT slices dynamically sampled into multi-crop global/local views & LeJEPA SSL + single-source (1S) dense auxiliary supervision using 118 TotalSegmentator classes (ReX inert) \\
\addlinespace
\textbf{DALE-CT-2S} & 2D ViT-Large (patch16) & CT-RATE cohort (25,692 unique scans) & 2D CT slices dynamically sampled into multi-crop global/local views & LeJEPA SSL + dual-source (2S) dense auxiliary supervision (TotalSegmentator and ReXGroundingCT) \\
\bottomrule
\end{tabular}%
}
\end{table*}

\section{Inference-Time Preprocessing Ablation}
\label{app:preproc}

The choice of preprocessing pipeline at inference time can meaningfully impact downstream performance for 2D slice-based models, as the encoder must process variable-resolution CT slices through a fixed-scale ViT. To systematically evaluate this, we ablated seven distinct strategies spanning three axes: body cropping (present vs.\ absent), input resolution (144, 256, or native up to 512 pixels), and tiling (single view vs.\ 5-cut). All experiments use the DALE-CT-2S backbone with a frozen encoder and a learned attention pooling head, selected via the same grid search protocol described in Section~\ref{sec:3d_probe}. To keep the embedding extraction tractable across all seven methods, we used a label-balanced subset of 5,000 patients from the CT-RATE training split.

\begin{table}[htbp]
\centering
\caption{Ablation of Inference-Time Preprocessing Strategies (DALE-CT-2S)}
\label{tab:preprocessing_ablation}
\resizebox{\columnwidth}{!}{%
\begin{tabular}{llcccc}
\toprule
\rowcolor{ukblue!10}\textbf{Resolution} & \textbf{Cropped} & \textbf{AUPRC} & \textbf{AUROC} & \textbf{Macro F1} & \textbf{BA} \\
\midrule
\multirow{2}{*}{144} & No & 0.5186 & 0.8183 & 0.5015 & 0.7045 \\
 & Yes & 0.5257 & 0.8156 & 0.5083 & 0.7067 \\
\addlinespace
\multirow{2}{*}{256} & No & 0.5382 & 0.8288 & 0.5209 & \textbf{0.7152} \\
 & Yes & \textbf{0.5421} & \textbf{0.8296} & 0.5137 & 0.7148 \\
\addlinespace
\multirow{2}{*}{Full $\leq512$} & No & 0.5324 & 0.8232 & 0.5216 & 0.7059 \\
 & Yes & 0.5315 & 0.8240 & 0.5221 & 0.7060 \\
\midrule
\multicolumn{6}{c}{\textbf{Alternative Tiling Strategy}} \\
\midrule
256 (Tiled 5-Cut) & Yes & 0.5391 & 0.8264 & \textbf{0.5252} & 0.7066 \\
\bottomrule
\end{tabular}%
}
\vspace{1ex}
\raggedright
\footnotesize \textit{Note: All methods use learned attention pooling (LR=0.01), which was universally optimal across all seven strategies.}
\end{table}

Table~\ref{tab:preprocessing_ablation} reports macro-averaged test metrics. Three findings emerge. First, body cropping consistently improves performance at matched resolutions: Cropped 256 outperforms Raw Resize 256 (0.5421 vs.\ 0.5382 AUPRC), and Cropped 144 outperforms Raw Resize 144 (0.5257 vs.\ 0.5186). By removing background pixels, body cropping ensures that nearly all patch tokens represent anatomical tissue. Second, performance increases from 144 pixels to 256 pixels resolution, then decreases slightly for full resolution (512 pixels). This reflects a pretraining mismatch: the DALE-CT encoder was trained exclusively with $256 \times 256$ and $144 \times 144$ crops, so at 512 pixels the ViT receives $4\times$ more patch tokens at a spatial scale it never encountered, requiring interpolation of learned positional embeddings; fine-tuning at full resolution could close this gap. Third, tiled inference offers a compelling middle ground: Tiled 5-Cut achieves the second-highest AUPRC and the highest Macro F1 (0.5252) by extracting five overlapping $256 \times 256$ body-cropped tiles, and concatenating their representations for each slice.

\section{Per-Class Benchmark Results}
\label{app:perclass}

Tables~\ref{tab:perclass_ctrate} and~\ref{tab:perclass_rad} tabulate per-class AUROC for every evaluated model, on CT-RATE and on RAD-ChestCT (frozen and retrained probe arms), under the head-to-head protocol of Section~\ref{sec:3d_probe}. Values are means over the available probe seeds (1--5 per model); the best model per row is bolded. DALE-CT models are abbreviated D-$\ast$; COLIPRI is COLIPRI-CRM.

\begin{table*}[tp]
\centering
\caption{Per-class AUROC on the CT-RATE test set ($n_{\text{test}}=992$).}
\label{tab:perclass_ctrate}
\scriptsize
\setlength{\tabcolsep}{4pt}
\begin{tabular}{lccccccccc}
\toprule
\rowcolor{ukblue!10}\textbf{Class} & \textbf{CT-CLIP} & \textbf{CT-FM} & \textbf{Merlin} & \textbf{COLIPRI} & \textbf{DINOv2} & \textbf{D-0} & \textbf{D-0-L} & \textbf{D-1S-v2} & \textbf{D-2S} \\
\midrule
Medical material & 0.636 & 0.796 & 0.821 & \textbf{0.929} & 0.840 & 0.829 & 0.827 & 0.817 & 0.876 \\
Arterial wall calcification & 0.659 & 0.877 & 0.900 & \textbf{0.926} & 0.901 & 0.914 & 0.912 & 0.912 & 0.915 \\
Cardiomegaly & 0.674 & 0.901 & 0.920 & 0.935 & \textbf{0.947} & 0.927 & 0.942 & 0.941 & 0.946 \\
Pericardial effusion & 0.616 & 0.772 & 0.839 & 0.833 & 0.815 & 0.834 & 0.866 & \textbf{0.867} & 0.858 \\
Coronary artery wall calcification & 0.639 & 0.878 & 0.886 & \textbf{0.935} & 0.895 & 0.899 & 0.905 & 0.899 & 0.898 \\
Hiatal hernia & 0.599 & 0.734 & 0.739 & 0.718 & 0.737 & 0.754 & 0.764 & 0.751 & \textbf{0.777} \\
Lymphadenopathy & 0.591 & 0.705 & 0.720 & \textbf{0.765} & 0.744 & 0.742 & 0.755 & 0.739 & 0.740 \\
Emphysema & 0.539 & 0.726 & 0.744 & 0.798 & 0.766 & 0.788 & 0.808 & 0.791 & \textbf{0.819} \\
Atelectasis & 0.609 & 0.717 & 0.732 & \textbf{0.805} & 0.711 & 0.730 & 0.757 & 0.755 & 0.763 \\
Lung nodule & 0.568 & 0.633 & 0.605 & \textbf{0.755} & 0.617 & 0.643 & 0.650 & 0.643 & 0.655 \\
Lung opacity & 0.565 & 0.692 & 0.713 & \textbf{0.881} & 0.745 & 0.791 & 0.788 & 0.772 & 0.840 \\
Pulmonary fibrotic sequela & 0.532 & 0.645 & 0.618 & \textbf{0.747} & 0.638 & 0.656 & 0.692 & 0.681 & 0.684 \\
Pleural effusion & 0.712 & 0.939 & 0.937 & \textbf{0.968} & 0.956 & 0.946 & 0.960 & 0.966 & 0.957 \\
Mosaic attenuation pattern & 0.612 & 0.824 & 0.821 & 0.894 & 0.846 & \textbf{0.896} & 0.889 & 0.873 & 0.881 \\
Peribronchial thickening & 0.580 & 0.756 & 0.780 & \textbf{0.813} & 0.801 & 0.780 & 0.787 & 0.780 & 0.786 \\
Consolidation & 0.603 & 0.750 & 0.767 & \textbf{0.932} & 0.800 & 0.809 & 0.804 & 0.806 & 0.855 \\
Bronchiectasis & 0.544 & 0.674 & 0.687 & \textbf{0.768} & 0.708 & 0.724 & 0.732 & 0.717 & 0.742 \\
Interlobular septal thickening & 0.593 & 0.787 & 0.829 & \textbf{0.855} & 0.815 & 0.842 & 0.836 & 0.851 & 0.852 \\
\bottomrule
\end{tabular}
\end{table*}

\begin{table*}[tp]
\centering
\caption{Per-class AUROC on the RAD-ChestCT test set ($n_{\text{test}}=360$): frozen probe (top block) and retrained probe (bottom block).}
\label{tab:perclass_rad}
\scriptsize
\setlength{\tabcolsep}{4pt}
\begin{tabular}{lccccccccc}
\toprule
\rowcolor{ukblue!10}\textbf{Class} & \textbf{CT-CLIP} & \textbf{CT-FM} & \textbf{Merlin} & \textbf{COLIPRI} & \textbf{DINOv2} & \textbf{D-0} & \textbf{D-0-L} & \textbf{D-1S-v2} & \textbf{D-2S} \\
\midrule
\multicolumn{10}{c}{\textbf{Frozen probe}} \\
\midrule
Calcification & 0.497 & 0.567 & 0.715 & \textbf{0.805} & 0.653 & 0.624 & 0.523 & 0.502 & 0.605 \\
Cardiomegaly & 0.539 & 0.525 & 0.864 & \textbf{0.888} & 0.817 & 0.834 & 0.846 & 0.810 & 0.870 \\
Pericardial effusion & 0.541 & 0.414 & 0.592 & 0.684 & 0.461 & 0.588 & 0.582 & \textbf{0.689} & 0.636 \\
Hiatal hernia & 0.508 & 0.452 & 0.651 & 0.690 & 0.634 & 0.534 & \textbf{0.732} & 0.641 & 0.519 \\
Emphysema & 0.447 & 0.526 & 0.700 & \textbf{0.842} & 0.660 & 0.628 & 0.767 & 0.729 & 0.645 \\
Atelectasis & 0.524 & 0.539 & 0.583 & \textbf{0.760} & 0.585 & 0.555 & 0.645 & 0.541 & 0.578 \\
Lung nodule & 0.487 & 0.637 & 0.668 & \textbf{0.743} & 0.696 & 0.534 & 0.627 & 0.657 & 0.624 \\
Lung opacity & 0.551 & 0.647 & 0.626 & \textbf{0.695} & 0.544 & 0.541 & 0.597 & 0.638 & 0.558 \\
Pulmonary fibrotic sequela & 0.429 & 0.439 & 0.512 & 0.540 & \textbf{0.608} & 0.514 & 0.456 & 0.363 & 0.521 \\
Pleural effusion & 0.560 & 0.547 & 0.828 & \textbf{0.932} & 0.742 & 0.745 & 0.780 & 0.799 & 0.719 \\
Peribronchial thickening & 0.534 & 0.582 & 0.593 & \textbf{0.632} & 0.539 & 0.425 & 0.485 & 0.466 & 0.524 \\
Consolidation & 0.518 & 0.724 & 0.681 & \textbf{0.827} & 0.588 & 0.654 & 0.578 & 0.709 & 0.634 \\
Bronchiectasis & 0.420 & 0.629 & 0.542 & \textbf{0.667} & 0.568 & 0.622 & 0.510 & 0.568 & 0.520 \\
Interlobular septal thickening & 0.561 & 0.743 & 0.655 & 0.747 & 0.732 & 0.626 & 0.732 & 0.765 & \textbf{0.769} \\
Lymphadenopathy & 0.576 & 0.533 & 0.549 & \textbf{0.666} & 0.544 & 0.556 & 0.640 & 0.628 & 0.619 \\
Medical material & 0.494 & 0.651 & 0.746 & \textbf{0.834} & 0.632 & 0.534 & 0.547 & 0.548 & 0.661 \\
\midrule
\multicolumn{10}{c}{\textbf{Retrained probe}} \\
\midrule
Calcification & 0.582 & 0.739 & 0.761 & \textbf{0.832} & 0.748 & 0.754 & 0.767 & 0.754 & 0.722 \\
Cardiomegaly & 0.628 & 0.865 & \textbf{0.922} & 0.902 & 0.917 & 0.882 & 0.916 & 0.884 & 0.871 \\
Pericardial effusion & 0.587 & 0.636 & 0.645 & 0.683 & 0.661 & \textbf{0.722} & 0.706 & 0.707 & 0.666 \\
Hiatal hernia & 0.529 & 0.623 & 0.634 & 0.675 & \textbf{0.746} & 0.737 & 0.695 & 0.699 & 0.744 \\
Emphysema & 0.638 & 0.819 & 0.760 & \textbf{0.888} & 0.834 & 0.849 & 0.885 & 0.839 & 0.830 \\
Atelectasis & 0.543 & 0.692 & 0.667 & \textbf{0.801} & 0.715 & 0.655 & 0.687 & 0.660 & 0.674 \\
Lung nodule & 0.621 & 0.714 & 0.680 & \textbf{0.786} & 0.669 & 0.684 & 0.707 & 0.707 & 0.696 \\
Lung opacity & 0.556 & 0.679 & 0.609 & \textbf{0.732} & 0.608 & 0.630 & 0.604 & 0.582 & 0.606 \\
Pulmonary fibrotic sequela & 0.614 & 0.798 & 0.803 & \textbf{0.852} & 0.850 & 0.832 & 0.847 & 0.813 & 0.829 \\
Pleural effusion & 0.681 & 0.886 & 0.874 & \textbf{0.942} & 0.907 & 0.901 & 0.892 & 0.895 & 0.917 \\
Peribronchial thickening & 0.572 & 0.596 & 0.600 & 0.655 & \textbf{0.717} & 0.683 & 0.711 & 0.658 & 0.701 \\
Consolidation & 0.478 & 0.708 & 0.711 & \textbf{0.850} & 0.705 & 0.702 & 0.691 & 0.668 & 0.698 \\
Bronchiectasis & 0.632 & 0.771 & 0.739 & 0.790 & \textbf{0.820} & 0.782 & 0.806 & 0.779 & 0.782 \\
Interlobular septal thickening & 0.547 & 0.682 & 0.712 & \textbf{0.787} & 0.767 & 0.744 & 0.779 & 0.727 & 0.689 \\
Lymphadenopathy & 0.464 & 0.733 & 0.651 & 0.725 & 0.681 & \textbf{0.739} & 0.722 & 0.691 & 0.711 \\
Medical material & 0.599 & 0.696 & 0.716 & \textbf{0.871} & 0.737 & 0.669 & 0.700 & 0.671 & 0.688 \\
\bottomrule
\end{tabular}
\end{table*}

\end{document}